\documentclass[lettersize,journal]{IEEEtran}
\usepackage{amsmath,amsfonts}

\usepackage{array}      
\usepackage{tabularx}   
\usepackage{booktabs}
\usepackage{multirow}
\usepackage{multicol}

\usepackage{makecell}      
\usepackage{xcolor}        
\usepackage{colortbl}      
\usepackage{threeparttable} 
\usepackage{graphicx}      

\usepackage{graphicx}
\usepackage{stfloats}
\usepackage[caption=false,font=normalsize,labelfont=sf,textfont=sf]{subfig}

\usepackage{algorithmic}
\usepackage{algorithm}

\usepackage{textcomp}
\usepackage{url}
\usepackage{verbatim}
\usepackage{cite}
\usepackage{pifont}    
\usepackage{utfsym}

\usepackage{hyperref}
\usepackage{cleveref}

\usepackage{xcolor}
\usepackage{soul}
\usepackage{cancel}

\UseRawInputEncoding

\usepackage{pdfpages} 
\usepackage{orcidlink}
\usepackage{amssymb}

\begin{document}


\title{Prior Evolution and Task Alignment \\ for Aerial Grasping}



\author{
Weiliang Deng$^{1}$,
Zhengyang Dang$^{1}$,
Yao Mu$^{2}$
and Ximin Lyu$^{1,3,\dagger}$
\thanks{${}^{\dagger}$ \textbf{Corresponding Author}.}
\thanks{%
${}^{1}$ School of Intelligent Systems Engineering,
Sun Yat-sen University, Guangzhou 510275, China
(e-mail: \texttt{\{dengwliang, dangzhy5\}@mail2.sysu.edu.cn};
\texttt{lvxm6@mail.sysu.edu.cn}).
}
\thanks{%
${}^{2}$ AI Institute, School of Computer Science,
Shanghai Jiao Tong University, Shanghai 200240, China
(e-mail: \texttt{muyao@sjtu.edu.cn}).
}
\thanks{%
${}^{3}$ Differential Robotics Technology Co., Ltd.,
Hangzhou, China.
}
}

\maketitle

\begin{abstract}
Aerial grasping is a remarkable capability exhibited by predatory birds, allowing them to capture prey through highly coordinated maneuvers in flight. 
Inspired by this capability, researchers have developed various formulations to reproduce such maneuvers through trajectory optimization. 
However, two limitations remain in practice.
First, the resulting optimization problem is highly nonconvex and sensitive to initialization, making high-quality solutions difficult to obtain under a limited computational budget. 
Second, prescribed numerical objectives are human-designed abstractions that describe successful grasping through a limited set of mathematically tractable quantities and may not fully capture what determines task success.
We investigate how learning can address these limitations within an analytical planner.
Accordingly, a trajectory prior is first learned from optimized motions and then evolved through a CEM-based process that evaluates sampled initializations with the deployed optimizer and retains favorable ones as new supervision. 
An Execution-Aware Critic learns from contact, lift, and completion outcomes to assess whether the optimized trajectories are likely to succeed in physical execution.
Its frozen energy can further serve as a differentiable grasping cost, allowing execution data to directly shape trajectory generation. 
Simulation and real-world experiments demonstrate improved optimization reliability, trajectory consistency, and grasping performance.
\end{abstract}



\section{Introduction}

\IEEEPARstart{A}{erial} manipulators combine the wide operational range of aerial robots with the ability to physically interact with objects and environments. These systems can operate in elevated, confined, hazardous, and otherwise difficult-to-access locations, enabling applications such as transportation, inspection, repair, construction, grasping, and perching \cite{ollero_past_2022,mohiuddin_survey_2020,zhang_aerial_2022,chermprayong_integrated_2019,hang_perching_2019,cao_proximal_2025,chen_ndob-based_2025,sun_agile_2025,zhang_an_ee_effector_oriented_2025,cao_eso-based_2024}. Among these capabilities, aerial grasping is particularly attractive because it enables a flying robot to acquire and retrieve an object without landing or establishing a stationary support. Inspired by predatory birds, researchers have reproduced such maneuvers through specialized mechanisms, dynamic modeling, control, and motion planning \cite{spica_aerial_2012,thomas_avian-inspired_2013,luo_time-optimal_2023,xu_biomimetic_2024,Ye_flyaware_2026}.

\begin{figure}[t]
    \centering
    \includegraphics[width=0.9\linewidth]{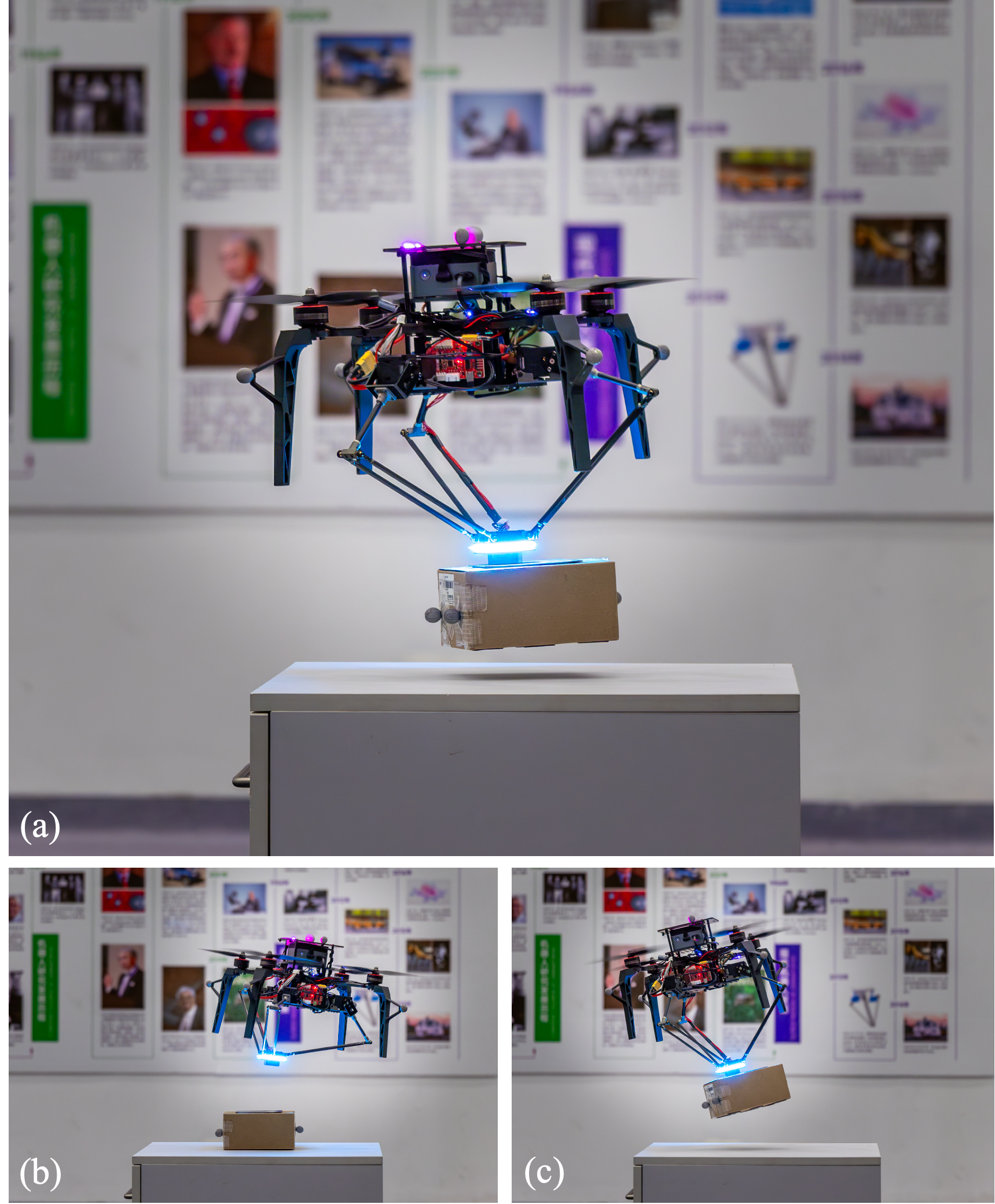}
    \caption{Snapshots of real-world aerial grasping. (a) The aerial manipulator holds the object after a successful grasp. (b) The delta arm extends as the end effector approaches the object. (c) Coordinated motion of the aerial platform and delta arm lifts the grasped object.}
    \label{fig:first_grasp}
\end{figure}

Effective aerial grasping requires coordinated motion planning for the aerial platform and manipulator, as shown in \Cref{fig:first_grasp}.
The planned trajectory must bring the end-effector to the object with an appropriate approach motion while satisfying geometric, temporal, kinematic and dynamic constraints \cite{chen_aerial_2025}.
To address these coupled requirements, differential-flatness methods exploit the structure of aerial manipulator dynamics, sampling-based methods search for feasible motions in complex spaces, and optimization methods provide a unified framework for incorporating geometric, dynamic, and task requirements \cite{welde_coordinate-free_2020,welde_dynamically_2021,kim_sampling-based_2019,tognon_control-aware_2018,luo_time-optimal_2023,cao_motion_2025}.
A previous work based on trajectory optimization developed a whole-body planning framework that jointly optimizes the aerial platform and manipulator trajectories and expresses manipulation tasks through flexible waypoint constraints, which serves as the analytical foundation of this work \cite{deng_wholebody_2025}.

Within such optimization-based planners, the intended grasping maneuver is typically specified through prescribed quantities such as target position, approach velocity, grasp orientation, and interception time \cite{spica_aerial_2012,thomas_avian-inspired_2013,luo_time-optimal_2023,chen_aerial_2025,deng_wholebody_2025}.
These quantities provide explicit and interpretable descriptions of the intended maneuver and can be optimized together with the known geometric and physical requirements.
However, these specifications are human-designed abstractions of successful grasping behavior, translating observed motion characteristics into quantities that can be explicitly prescribed in optimization.
This leaves two distinct challenges:
finding favorable solutions to the resulting nonconvex optimization problem under a limited computational budget, and accounting for the mismatch between numerical objectives and physical grasping success.
These challenges motivate the central question of this work: how can learning address these limitations within trajectory optimization?

The first challenge concerns the reliable solution of the nonconvex trajectory optimization problem.
The coupling of the prescribed grasping requirements with geometric, temporal, kinematic and dynamic constraints produces a highly nonconvex problem with multiple local solutions.
Local trajectory optimizers can efficiently refine an initial trajectory, but both their convergence behavior and the final solution depend strongly on the initialization \cite{Schulman_finding_2013,kang_fast_2024,joao_motion_2025}. 
A poorly structured initialization can lead to an unfavorable local solution or require substantial solver refinement, the latter being particularly problematic when planning must be completed under a limited computational budget.
Sampling- and search-based methods can provide collision-free reference paths, but their outputs often require further smoothing and dynamic refinement, and generating the initial path can itself require considerable computation \cite{kim_sampling-based_2019,Kalakrishnan_stomp_2011,Lembono_memory_2020}. 
Initialization therefore affects not only how much refinement is required, but also which local solution the optimizer ultimately reaches.

Trajectory priors provide a promising way to reuse motion structures from previously solved problems and generate informative initializations for local optimization \cite{Lampariello_trajectory_2011,Mansard_using_2018,Lembono_memory_2020,joao_motion_2025}.
Existing approaches store previous solutions, learn mappings from task conditions to trajectories, learn sampling distributions or implicit trajectory priors, or learn multimodal trajectory distributions with generative models \cite{ichter_learning_2018,Dragan_learning_2017,Urain_learning_2022,Huang_diffusionseeder_2024,Haffemayer_warm-starting_2026}.
These studies show that learning an informative trajectory prior from previously solved problems is a common approach for improving the initialization of subsequent motion optimization. 
However, most learned trajectory priors are trained from fixed offline datasets to reproduce previously optimized trajectories or their distributions. 
This objective assumes that similarity to an offline solution is a useful proxy for initialization quality.
Yet these two notions need not coincide: an initialization that closely matches an offline trajectory may still require substantial solver refinement, whereas a different initialization may lead the optimizer to a more favorable local solution \cite{Dragan_learning_2017,Lembono_memory_2020}.
This distinction motivates our first investigation: whether a trajectory prior should merely reproduce previously optimized solutions or continue to adapt according to the behavior of the deployed optimizer.

To investigate this possibility, we need a mechanism that can expose the prior to alternative initializations and evaluate them through the deployed optimizer.
Stochastic optimization naturally provides such a mechanism.
Existing methods perturb candidate trajectories and evaluate the resulting solutions during optimization, but the information discovered through this exploration is generally discarded once the current planning problem is solved \cite{Kalakrishnan_stomp_2011}.
Experience-based approaches retain newly optimized trajectories for subsequent queries, yet relying on an expanding trajectory database becomes increasingly difficult in high-dimensional spaces \cite{Tseng_adaptive_2024}.
This suggests a different use of repeated optimization: the experience accumulated by the optimizer can be transferred into a shared parametric prior.
We investigate this idea using the sampling-and-selection principle of the cross-entropy method (CEM) \cite{deboer_a_tutorial_2003}.
Around the current prior prediction, alternative initializations are sampled and independently refined by the deployed trajectory optimizer.
The selected optimizer inputs are then retained as new supervision for the shared neural prior.
Repeating this process allows the prior to evolve according to the behavior of the deployed optimizer, rather than merely reproducing the solutions contained in a fixed offline dataset.

The second challenge concerns the gap between analytical optimality and physical task success.
Prescribed grasping requirements provide explicit and interpretable descriptions of the intended maneuver, but they encode successful grasping through human-designed motion specifications.
Physical outcomes can additionally depend on contact quality, effective grasp depth, object disturbance, retention after pickup, and other interaction effects that are difficult to specify analytically \cite{song_task-based_2015,hadfieldmenell_inverse_2017}.
A trajectory can therefore satisfy the prescribed requirements and attain a low numerical objective while still failing during physical execution.
This discrepancy suggests that solving the prescribed optimization problem more effectively is insufficient when the objective itself does not fully reflect what makes a grasp succeed.

Learning from interaction offers a way to recover task information that is difficult to prescribe analytically \cite{song_reaching_2023,Kun_performant_2026}.
Reinforcement learning and policy learning have demonstrated increasingly capable robotic manipulation and have also been applied to aerial manipulation systems \cite{cuniato_learning_2023,dimming_non_prehensile_2024,he_flying_2025}.
However, many requirements in aerial grasping, including geometric, temporal, kinematic and dynamic constraints, are already represented explicitly by the analytical planner and need not be relearned from execution data \cite{Wei_SafeDiffuser_2023,Haffemayer_warm-starting_2026}.
Sparse physical success signals can also make direct policy learning expensive and data intensive \cite{saxena_deep_2026,wu_precise_2026}.
This motivates retaining trajectory optimization as the planning backbone and using execution outcomes to complement what the analytical formulation does not capture.
A key design question is where such information should enter the planning pipeline.
In the main framework, we introduce execution supervision after trajectory optimization, where it evaluates complete solutions without modifying the analytical trajectory optimization process.
We later examine this design experimentally, comparing alternative placements of execution supervision and investigating whether the learned execution preference can also play a more direct role in trajectory optimization.

These two limitations expose two distinct roles for learning.
Repeated optimization provides information about which initializations lead the deployed solver toward favorable local solutions with limited refinement, whereas physical execution reveals which motion characteristics matter for successful task completion.

In this work, we investigate these two roles within an optimization-based framework for aerial grasping.
We first analyze how the prescribed grasping requirements shape the nonconvex optimization landscape and lead to a misalignment between numerical objective values and physical task success.
We then learn a trajectory prior from optimized motions and further evolve it using the deployed optimizer, allowing repeated optimization experience to improve future initialization.
An Execution-Aware Critic (EAC) learns from contact, lift, and completion outcomes to assess complete optimized trajectories without replacing the analytical planner.
Through paired simulation and real-world experiments, we examine what is learned from optimization and execution, where execution supervision should enter the planning process, and how these choices affect trajectory consistency and physical grasping performance.
Most notably, we find that the learned execution preference can also serve as a differentiable grasping cost, allowing physical execution outcomes to directly shape trajectory optimization.

The main contributions of this work are as follows.

1) \textbf{Understanding the Limitations of Optimization-based Aerial Grasping:} We analyze how the prescribed grasp position, velocity, and orientation penalties interact with temporal, geometric, kinematic and dynamic requirements. 
We show that their interaction produces distinct local solution modes and makes the resulting optimization sensitive to trajectory initialization.
We further identify objective-task misalignment: trajectories can achieve low analytical costs yet still fail during physical execution.

2) \textbf{Learning Trajectory Initialization from the Deployed Optimizer:} We introduce a trajectory prior that first captures recurring motion structures from previously optimized trajectories and then evolves through CEM-based sampling, optimization, selection, and replay. This process adapts the prior according to which predictions actually provide effective initializations for the deployed optimizer.
Beyond improving optimization reliability, we show that the evolved prior produces a more consistent family of grasping trajectories under initialization perturbations and varying task conditions.

3) \textbf{Learning from Physical Execution for Trajectory Evaluation and Optimization:}
We introduce an Execution-Aware Critic learned from contact, lift, and completion outcomes to assess complete optimized trajectories.
Through controlled comparisons, we examine where execution supervision should enter the planning process and show that using it to evaluate complete optimized trajectories provides clear improvements without modifying the evolved trajectory prior.
More importantly, the same frozen critic can serve as a differentiable grasping cost, allowing execution outcomes to directly shape trajectory optimization while retaining the analytical treatment of the remaining planning requirements.

The remainder of this paper is organized as follows. \Cref{sec:grasp_formulation} presents the aerial grasping formulation and analyzes its nonconvex solution modes and the misalignment between analytical cost and physical task success. \Cref{sec:self_evolving_prior} introduces the trajectory prior, Offline Supervised Learning, CEM-based Online Evolution, EAC, and the deployment procedure. \Cref{sec:experiments} presents the simulation and real-world experiments, while \Cref{sec:discussion} discusses the extensibility. Finally, \Cref{sec:conclusion} concludes the paper.

\section{Aerial Grasping Formulation and Analysis}
\label{sec:grasp_formulation}

\begin{figure}[t]
    \centering
    \includegraphics[width=0.8\linewidth]{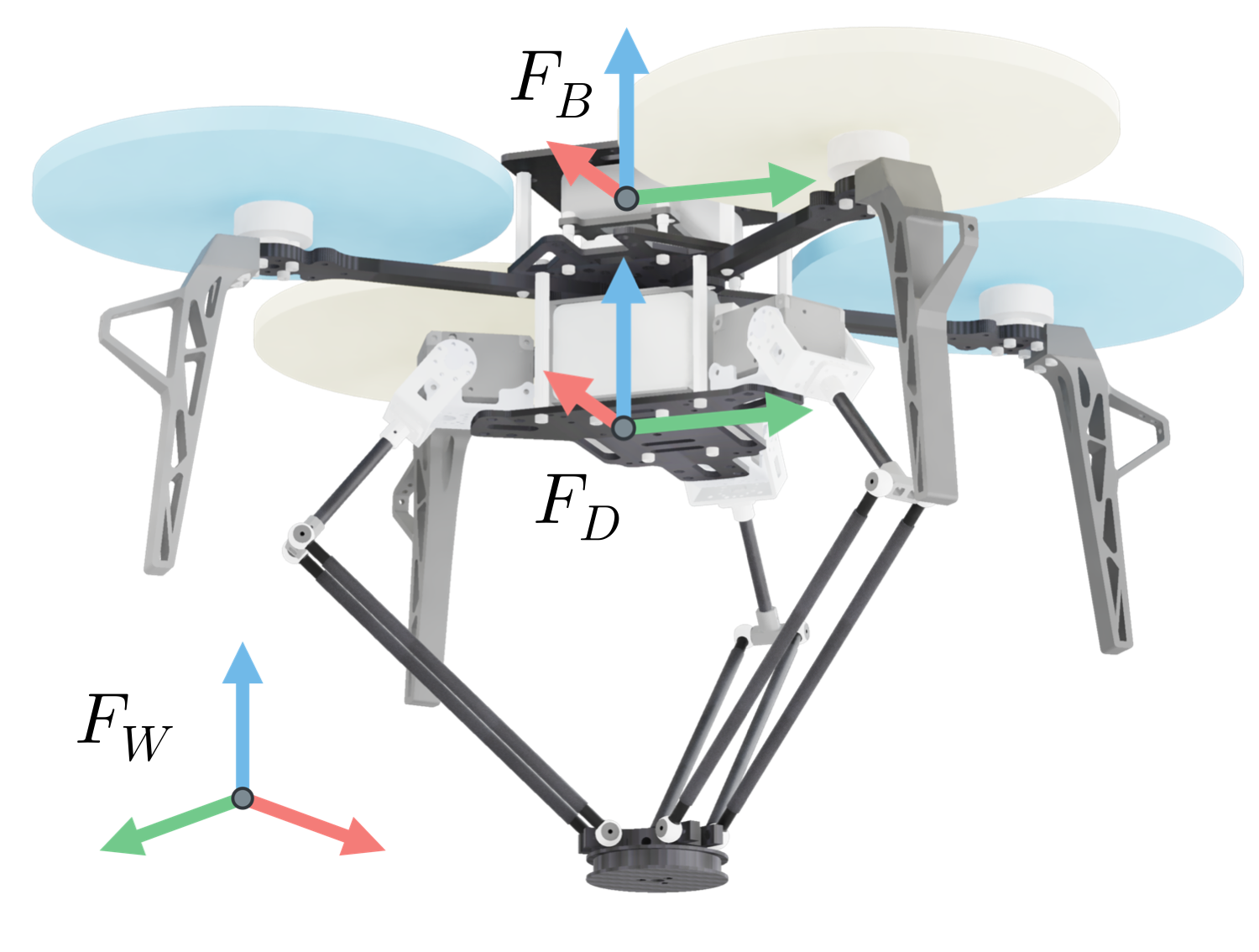}
    \caption{Coordinate frames of the aerial manipulator. The world frame $F_W$, body frame $F_B$, and delta frame $F_D$ are defined at the inertial reference, the nominal center of mass of the quadrotor, and the center of the static platform of the delta arm, respectively.}
    \label{fig:system_setup}
\end{figure}

\subsection{System Setup}
\label{subsec:system_setup}

We consider a quadrotor-based aerial manipulator equipped with a 3-DoF parallel delta arm and a grasping end-effector, as illustrated in \Cref{fig:system_setup}. Three coordinate frames are defined: the world frame $F_W=\{\boldsymbol{x}_W,\boldsymbol{y}_W,\boldsymbol{z}_W\}$, the body frame $F_B=\{\boldsymbol{x}_B,\boldsymbol{y}_B,\boldsymbol{z}_B\}$ located at the nominal center of mass of the quadrotor, and the delta frame $F_D=\{\boldsymbol{x}_D,\boldsymbol{y}_D,\boldsymbol{z}_D\}$ fixed to the center of the static platform of the delta arm. The frames $F_B$ and $F_D$ differ only by a fixed translation.

Unless otherwise specified, vectors are expressed in $F_W$, while a left superscript indicates the corresponding reference frame. Let $\boldsymbol{p}_b$ and $\boldsymbol{v}_b$ denote the position and velocity of the quadrotor in $F_W$, and let ${}^{D}\boldsymbol{p}_e$ denote the end-effector position in $F_D$. The corresponding end-effector position in the world frame is given by
\begin{equation}
\boldsymbol{p}_e
=
\mathbf{R}_B
\left(
{}^{B}\boldsymbol{p}_D
+
{}^{D}\boldsymbol{p}_e
\right)
+
\boldsymbol{p}_b,
\label{eq:ee_world_position}
\end{equation}
where $\mathbf{R}_B\in SO(3)$ denotes the rotation from $F_B$ to $F_W$, and ${}^{B}\boldsymbol{p}_D$ is the fixed translation from $F_B$ to $F_D$. The world-frame end-effector velocity is denoted by $\boldsymbol{v}_e=\dot{\boldsymbol{p}}_e$ and is obtained by differentiating \Cref{eq:ee_world_position}.

\subsection{Trajectory Optimization Formulation}
\label{subsec:spatiotemporal_formulation}

We adopt a hybrid Cartesian representation in which the quadrotor trajectory is expressed in $F_W$, while the end-effector motion relative to the platform is expressed in $F_D$. The resulting 6-D whole-body trajectory is defined as
\begin{equation}
\boldsymbol{p}(t)
=
\begin{bmatrix}
\boldsymbol{p}_b^{\top}(t) &
{}^{D}\boldsymbol{p}_e^{\top}(t)
\end{bmatrix}^{\top}
\in\mathbb{R}^{6}.
\label{eq:whole_body_trajectory}
\end{equation}

Following the spatio-temporal parameterization in \cite{wang_geometrically_2022}, the polynomial trajectory is determined by its intermediate whole-body waypoints and segment durations. Specifically, the trajectory contains $M-1$ intermediate whole-body waypoints and $M$ segment durations. Each intermediate waypoint is defined as
\begin{equation}
\boldsymbol{w}_i
=
\begin{bmatrix}
\boldsymbol{p}_{b,i}^{\top} &
{}^{D}\boldsymbol{p}_{e,i}^{\top}
\end{bmatrix}^{\top}
\in\mathbb{R}^{6},
\qquad
i\in\{1,\ldots,M-1\}.
\label{eq:whole_body_waypoint}
\end{equation}
The resulting compact spatio-temporal representation is
\begin{equation}
\boldsymbol{z}
=
\begin{bmatrix}
\boldsymbol{w}_1^{\top} &
\cdots &
\boldsymbol{w}_{M-1}^{\top} &
T_1 &
\cdots &
T_M
\end{bmatrix}^{\top}
\in\mathbb{R}^{7M-6},
\label{eq:trajectory_parameter}
\end{equation}
where $T_i$ denotes the duration of the $i$-th trajectory segment.

Using this representation, aerial grasping is formulated as the following nonlinear trajectory optimization problem:
\begin{subequations}
\label{eq:aerial_grasping_problem}
\begin{align}
\underset{\boldsymbol{z}}{\min}\quad
\mathcal{J}
&=
\int_{0}^{T_{\sigma}}
\left\|
\boldsymbol{p}^{(s)}(t)
\right\|^{2}
\,\mathrm{d}t
+
\rho T_{\sigma},
\label{eq:grasping_objective}
\\
\mathrm{s.t.}\quad
&
\mathcal{G}
\left(
\boldsymbol{p}(t),
\ldots,
\boldsymbol{p}^{(s)}(t)
\right)
\preceq
\boldsymbol{0},
\qquad
\forall t\in[0,T_{\sigma}],
\label{eq:grasping_constraints}
\\
&
\boldsymbol{p}^{[s-1]}_1(0)
=
\boldsymbol{p}_o,
\qquad
\boldsymbol{p}^{[s-1]}_M(T_M)
=
\boldsymbol{p}_f,
\label{eq:grasping_boundary_conditions}
\\
&
T_i>0,
\qquad
i\in\{1,\ldots,M\}.
\label{eq:grasping_positive_time}
\end{align}
\end{subequations}
where $T_{\sigma}=\sum_{i=1}^{M}T_i$ is the total trajectory duration, $\rho>0$ balances control effort and maneuver time, and $\boldsymbol{p}^{[s-1]}=[\boldsymbol{p},\dot{\boldsymbol{p}},\ldots,\boldsymbol{p}^{(s-1)}]$ collects the trajectory derivatives. The vectors $\boldsymbol{p}_o$ and $\boldsymbol{p}_f$ denote the prescribed initial and terminal conditions, while $\mathcal{G}$ collects the geometric, kinematic and dynamic constraints \cite{yang_whole-body_2021,ren_online_2023,ren_safety_2025}.

To define the grasping behavior, the most fundamental requirement is to ensure that the end effector reaches the desired position. This requirement is commonly incorporated into trajectory optimization as an end-effector position penalty \cite{luo_time-optimal_2023,chen_aerial_2025,spica_aerial_2012}. The end-effector position penalty is defined as
\begin{equation}
J_{pe}
=
\left\|
\boldsymbol{p}_{e,t_g}
-
\boldsymbol{p}^{\mathrm{des}}_{e}
\right\|^{2},
\label{eq:grasp_position_penalty}
\end{equation}
where $\boldsymbol{p}_{e,t_g}$ is the end-effector position expressed in $F_W$ at the specific moment $t_g$, and $\boldsymbol{p}^{\mathrm{des}}_{e}$ is the desired grasp position.

Reaching the desired position alone does not sufficiently specify the grasping motion. Therefore, the end-effector velocity at the grasp moment is further regulated through the velocity penalty. In \cite{luo_time-optimal_2023}, the end-effector velocity is required to remain close to the target velocity. Since the target is stationary in the considered scenario, the desired velocity can be set to zero. Alternatively, \cite{deng_wholebody_2025} assigns a small downward velocity to the end effector to promote contact. The end-effector velocity penalty is defined as
\begin{equation}
J_{ve}
=
\left\|
\mathbf{S}_v
\left(
\boldsymbol{v}_{e,t_g}
-
\boldsymbol{v}^{\mathrm{des}}_{e}
\right)
\right\|^{2},
\label{eq:grasp_velocity_penalty}
\end{equation}
where $\mathbf{S}_v\in\mathbb{R}^{3\times3}$ is a diagonal selection matrix and $\boldsymbol{v}^{\mathrm{des}}_{e}$ is the desired end-effector velocity. 

For grasping tasks with an orientation requirement, an additional orientation penalty is imposed at the grasp moment \cite{chen_aerial_2025,deng_wholebody_2025}. This requirement is expressed through the thrust direction because the delta-arm end-effector shares the orientation of the aerial platform:
\begin{equation}
J_{oe}
=
\left\|
\boldsymbol{f}_{t_g}
-
f
\boldsymbol{o}^{\mathrm{des}}
\right\|^{2},
\label{eq:grasp_orientation_penalty}
\end{equation}
where $\boldsymbol{o}^{\mathrm{des}}$ denotes the desired unit thrust direction and
$f=\|\boldsymbol{f}_{t_g}\|$ denotes the thrust magnitude at the grasp waypoint.

Following the implementation in \cite{deng_wholebody_2025}, the task-relative waypoint, velocity, and orientation constraints are incorporated into \Cref{eq:grasping_objective}.

Finally, the continuously evaluated penalty terms are approximated using fixed numerical samples along each polynomial segment \cite{wang_geometrically_2022}:
\begin{equation}
\int_{0}^{T_{\sigma}}
J_{\mathrm{c}}(t)\,\mathrm{d}t
\approx
\sum_{i=1}^{M}
\sum_{n=0}^{N-1}
J_{\mathrm{c}}^{i}(t_{i,n})
\Delta t = \hat{\mathcal{J}},
\label{eq:sampled_penalty_evaluation}
\end{equation}
where $J_{\mathrm{c}}$ denotes the continuously evaluated penalty terms and $N$ is the number of samples per segment. Together with the grasp-related penalties and the remaining objective terms, these sampled penalties form the complete numerical objective $\widehat{\mathcal{J}}$ evaluated by the trajectory optimizer.

\subsection{Nonconvexity and Local Minima under Progressive Grasp Requirements}
\label{subsec:constraint_shaping}

\begin{figure}[t]
    \centering
    \includegraphics[width=\linewidth]{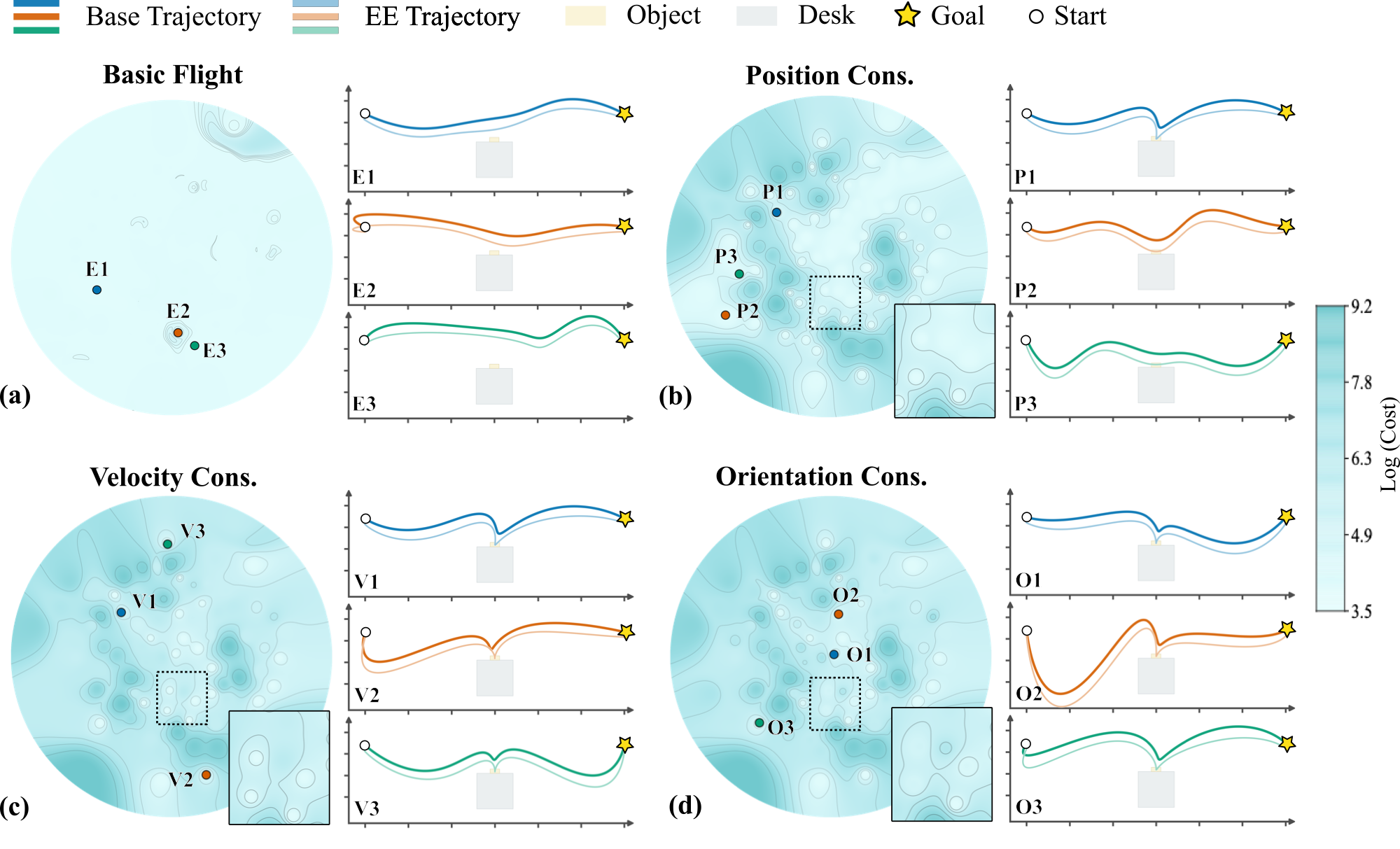}
    \caption{Evolution of the trajectory-cost landscape under progressively introduced grasping penalties. From (a) to (d), the end-effector position, velocity, and orientation penalties are progressively added to the basic flight objective. In each panel, the left plot shows the cost landscape, while the right plots show representative aerial platform and end-effector trajectories corresponding to the marked points.}
    \label{fig:constraint_shaping}
\end{figure}

The optimization problem in \Cref{eq:aerial_grasping_problem} is highly nonconvex and sensitive to trajectory initialization \cite{joao_motion_2025}. Even without grasp-specific requirements, multiple geometrically distinct collision-free trajectories can satisfy the same boundary conditions. Consequently, different initializations may lead the local optimizer to substantially different solutions under identical planning conditions.

To examine how grasp-specific requirements affect this initialization sensitivity, we adopt the formulation in \cite{deng_wholebody_2025} as a representative example because it incorporates comprehensive whole-body grasping requirements. We use the basic flight objective as the baseline and progressively introduce the end-effector position, velocity, and orientation penalties at the prescribed grasp waypoint:
\begin{subequations}
\label{eq:progressive_objectives}
\begin{align}
\widehat{\mathcal{J}}^{(1)}(\boldsymbol{z})
&=
\widehat{\mathcal{J}}^{(0)}(\boldsymbol{z})
+
\lambda_e J_{pe}(\boldsymbol{z}),
\\
\widehat{\mathcal{J}}^{(2)}(\boldsymbol{z})
&=
\widehat{\mathcal{J}}^{(1)}(\boldsymbol{z})
+
\lambda_v J_{ve}(\boldsymbol{z}),
\\
\widehat{\mathcal{J}}^{(3)}(\boldsymbol{z})
&=
\widehat{\mathcal{J}}^{(2)}(\boldsymbol{z})
+
\lambda_o J_{oe}(\boldsymbol{z}),
\end{align}
\end{subequations}
where $\widehat{\mathcal{J}}^{(0)}$ denotes the basic flight objective without grasp-specific terms.

To visualize how these additional grasping requirements reshape the optimization landscape, we fix the boundary conditions, scene geometry, trajectory representation, solver settings, and non-grasping penalty terms, and optimize a large set of randomly initialized trajectories under each formulation in \Cref{eq:progressive_objectives}.
Each initialization is refined independently by the same local optimizer, and the converged trajectory together with its final objective value is recorded.
In each panel of \Cref{fig:constraint_shaping}, the left plot shows the trajectory-cost landscape constructed by interpolation among the locally optimized trajectories, with the marked points identifying representative local solutions. The corresponding aerial platform and end-effector trajectories are shown on the right.

As shown in \Cref{fig:constraint_shaping} (a), the base formulation already admits multiple local optima. Nevertheless, the cost landscape remains relatively flat, with broad low-cost regions. A wide range of initializations can therefore converge to solutions with similar objective values. After the end-effector position penalty is introduced in \Cref{fig:constraint_shaping} (b), the cost landscape becomes substantially more complex. High-cost regions separate the previously broad low-cost area into several distinct regions associated with different local optima, making the optimization more sensitive to initialization.
When the end-effector velocity penalty is further introduced in \Cref{fig:constraint_shaping} (c), the overall landscape remains similar to that in \Cref{fig:constraint_shaping} (b), but some previously low-cost local optima become unfavorable under the additional velocity requirement. Since the local optimizer can only converge within the basin determined by its initialization, an initialization leading to one of these unfavorable local optima may no longer yield a usable grasping trajectory.
For grasping tasks with a prescribed orientation, the position and velocity requirements in \Cref{fig:constraint_shaping} (c) remain insufficient. When the end-effector orientation penalty is further introduced in \Cref{fig:constraint_shaping} (d), more of the previously favorable local optima become unfavorable, further reducing the set of local solutions that satisfy the grasping requirements. 
At the same time, the representative trajectories show that the aerial platform adjusts its attitude near the grasp waypoint to better satisfy the desired end-effector orientation.

These results show that progressively introducing grasp-specific requirements reduces the set of favorable local optima, making the choice of initialization increasingly important. Because the optimizer is local, its final solution depends not only on the objective itself but also on the basin of attraction in which the initialization  lies\cite{kang_fast_2024}. This observation directly motivates the use of an informative trajectory prior: its role is not merely to accelerate optimization, but more importantly to initialize the solver near a favorable grasping mode and thereby reduce convergence to poor local minima.

\subsection{Emergent Avian-Like Motion and Objective-Task Misalignment}
\label{subsec:motion_ambiguity}

The grasp-related penalties specify the desired end-effector state at the designated grasp moment, but they do not explicitly prescribe the complete motion before and after contact. 
Nevertheless, an avian-like grasping maneuver can emerge from the coupling between the temporal objective and the whole-body trajectory optimization. 
At the grasp waypoint, the world-frame end-effector velocity can be decomposed as
\begin{equation}
\boldsymbol{v}_{e,t_g}
=
\boldsymbol{v}_{b,t_g}
+
\boldsymbol{v}_{\mathrm{arm},t_g}
+
\boldsymbol{v}_{\mathrm{rot},t_g},
\label{eq:grasp_velocity_decomposition}
\end{equation}
where $\boldsymbol{v}_{b,t_g}$ is the translational velocity of the aerial platform, $\boldsymbol{v}_{\mathrm{arm},t_g}$ is induced by the relative motion of the manipulator, and $\boldsymbol{v}_{\mathrm{rot},t_g}$ is induced by the aerial platform rotation.

\begin{figure}[t]
    \centering
    \includegraphics[width=\linewidth]{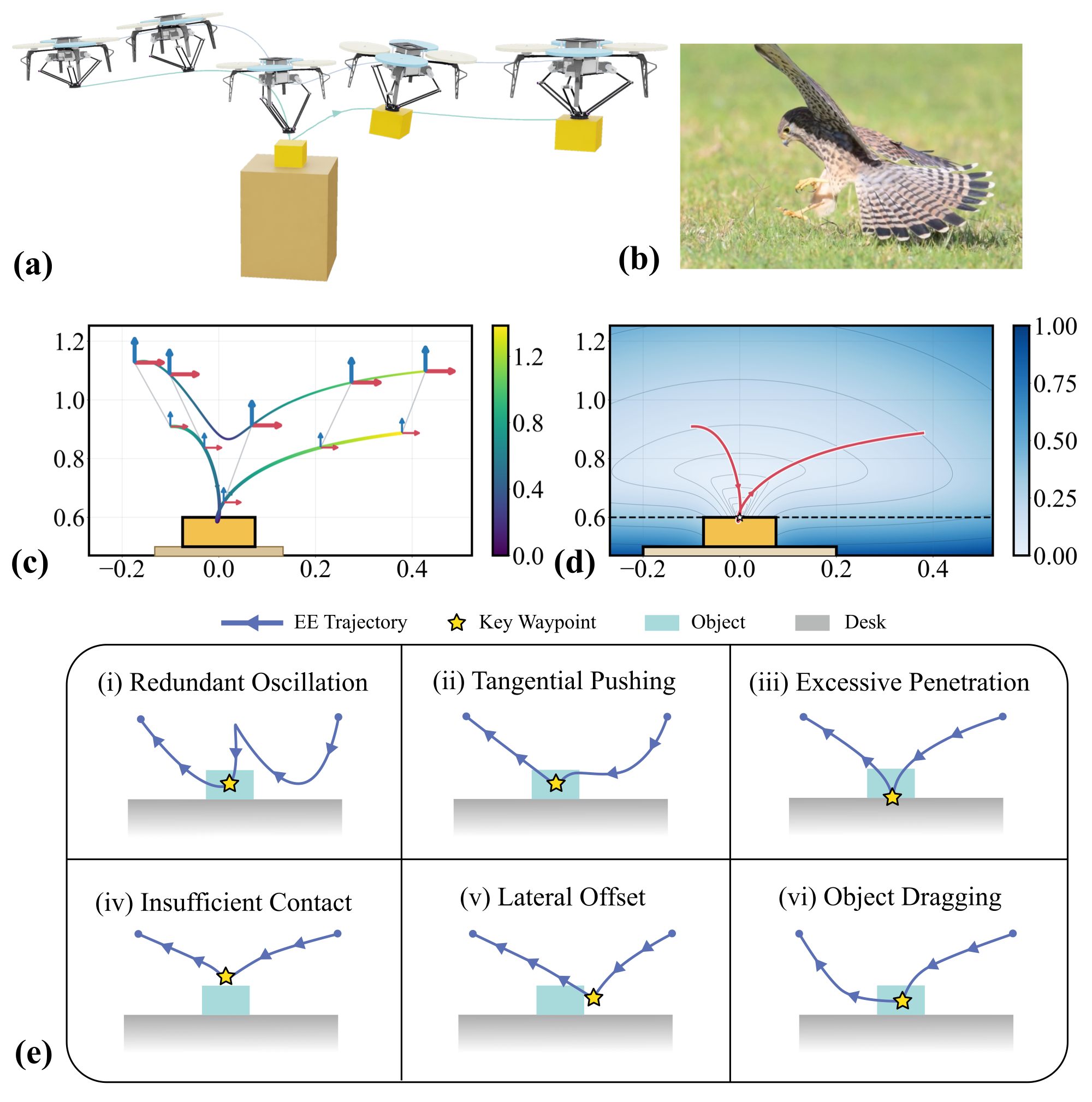}
    \caption{Emergent avian-like motion and objective-task misalignment in aerial grasping. (a)-(d) The optimized whole-body motion generates a backward end-effector sweep similar to avian prey capture. (e) Representative low-cost trajectories that remain undesirable for physical execution.}
    \label{fig:motion_ambiguity}
\end{figure}

The temporal objective favors completing the maneuver within a short duration rather than bringing the aerial platform to a full stop, whereas the end-effector velocity penalty regulates the selected velocity components toward their desired values.
When the platform retains a nonzero velocity at this waypoint, the arm motion and platform rotation must compensate for it:
\begin{equation}
\mathbf{S}_v
\left(
\boldsymbol{v}_{\mathrm{arm},t_g}
+
\boldsymbol{v}_{\mathrm{rot},t_g}
\right)
\approx
\mathbf{S}_v
\left(
\boldsymbol{v}^{\mathrm{des}}_{e,t_g}
-
\boldsymbol{v}_{b,t_g}
\right).
\label{eq:grasp_velocity_compensation}
\end{equation}
As illustrated in \Cref{fig:motion_ambiguity} (a)-(c), this compensation can cause the end effector to extend outward during the approach and sweep backward relative to the platform around the grasping instant. This motion resembles the avian capture strategy shown in \Cref{fig:motion_ambiguity} (b), in which the legs and claws sweep backward to reduce their relative velocity with respect to the target
\cite{thomas_avian-inspired_2013,luo_time-optimal_2023}.

As shown in \Cref{fig:motion_ambiguity} (d), the optimized trajectory approaches the object from above, reaches the designated grasp waypoint, and then departs upward. 
The end-effector position and velocity penalties encourage the end effector to attain the prescribed position and velocity at $t_g$, while the collision penalties make trajectories passing through the object or supporting surface costly. 
Together with the smoothness and temporal terms, these penalties can result in a V-shaped trajectory, although this shape is not explicitly prescribed.

However, because the end-effector position, velocity, and orientation requirements are imposed as weighted penalties together with the other terms in the objective, the optimizer may trade one requirement against another. 
With an unfavorable initialization, it may converge to a local solution that lowers the total cost by accepting small residual errors in some requirements while better satisfying others. 
As illustrated in \Cref{fig:motion_ambiguity} (e), such trajectories can therefore achieve a low overall objective value while still exhibiting unnecessary oscillation, lateral offset, insufficient contact, excessive penetration, or object dragging.

We refer to the discrepancy between the behavior favored by the numerical objective and that required by the physical task as \emph{objective-task misalignment}. This is different from the nonconvexity discussed in the previous section. Nonconvexity causes different initializations to converge to different local optima, whereas objective-task misalignment means that a trajectory with a low objective value may still perform poorly in the actual grasping task. Therefore, an informative trajectory prior is required not only to guide the optimizer toward a favorable local optimum, but also to produce a trajectory that is more suitable for physical execution. Since the numerical objective does not fully reflect the quality of physical contact, execution feedback is further required to distinguish trajectories according to their actual grasping performance.

\begin{figure*}[t]
    \centering
    \includegraphics[width=\textwidth]{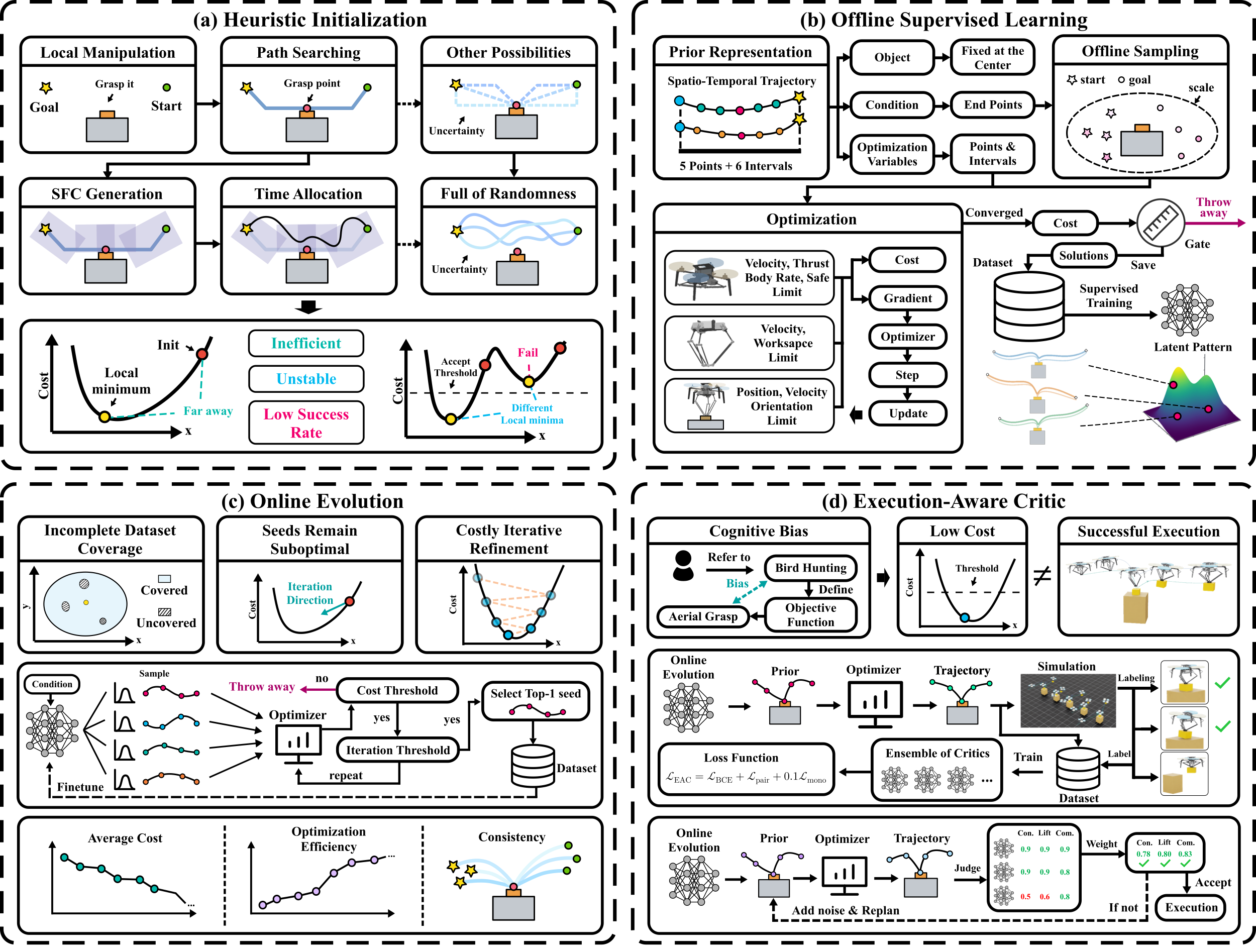}
    \caption{Overview of the proposed framework.
(a) Heuristic Initialization generates an initial trajectory using path search and SFC construction before trajectory optimization.
(b) Offline Supervised Learning trains a deterministic prior from optimizer-refined trajectories to predict trajectory initializations.
(c) Online Evolution samples neighboring initializations around the prior prediction, evaluates them using the deployed optimizer, and updates the prior with selected optimizer inputs.
(d) Execution-Aware Critic learns from physical execution outcomes and evaluates optimized trajectories to provide execution-aware assessment during deployment.}
    \label{fig:framework_overview}
\end{figure*}

\section{Prior Evolution and Task Alignment}
\label{sec:self_evolving_prior}

\subsection{Framework Overview}
\label{subsec:framework_overview}

The trajectory planner introduced in the preceding section explicitly represents geometric, temporal, kinematic, dynamic, and grasping requirements. 
Nevertheless, its performance remains limited by two factors: the sensitivity of nonconvex optimization to trajectory initialization and the discrepancy between numerical objective values and physical execution outcomes.

\Cref{fig:framework_overview} summarizes the whole framework, which incorporates optimizer feedback and execution feedback at different stages of trajectory generation. Heuristic Initialization first provides an initial trajectory for the optimizer through geometric path search and Safe Flight Corridor (SFC) construction. Offline Supervised Learning then captures recurring whole-body trajectory structures from optimized solutions and learns a deterministic prior that predicts optimizer initializations from task conditions. Starting from this offline prior, Online Evolution applies a CEM-based sampling and selection process around the current prediction, evaluates candidate initializations using the deployed optimizer, and uses the selected optimizer inputs as new supervision to adapt the prior toward more favorable optimization regions.

After the prior is frozen, the Execution-Aware Critic introduces physical execution feedback into the planning pipeline. It learns from contact, lift, and task completion outcomes to evaluate complete optimized trajectories. During deployment, the prior generates an initialization, the trajectory optimizer refines the motion while preserving analytical constraints, and EAC determines whether the resulting trajectory satisfies the expected execution quality or whether another initialization should be generated within the available computational budget.

\subsection{Trajectory Prior Representation}
\label{subsec:prior_representation}
An aerial grasping maneuver has a strong local structure around the grasp point. 
Motions performed at different world locations can therefore be represented in a common local coordinate frame centered at the corresponding grasp point.
We train the trajectory prior by sampling the boundary conditions within a bounded region of this local frame, allowing the learned motion structure to be reused at different grasp locations without repeatedly learning equivalent world-frame translations.

The prior input consists of the initial and terminal platform positions expressed in the local frame:
\begin{equation}
\boldsymbol{c}
=
\begin{bmatrix}
\widetilde{\boldsymbol{p}}_{b,o}^{\top} &
\widetilde{\boldsymbol{p}}_{b,f}^{\top}
\end{bmatrix}^{\top}
\in\mathbb{R}^{6},
\label{eq:local_prior_condition}
\end{equation}
where $\widetilde{\boldsymbol{p}}_{b,o}$ and
$\widetilde{\boldsymbol{p}}_{b,f}$ denote the local initial and terminal
platform positions, respectively. During data generation, these positions are
sampled within the prescribed local range.

A fixed trajectory dimension is required for prior learning. We use six
trajectory segments and five intermediate whole-body waypoints, following the
representation in \Cref{eq:trajectory_parameter}. The initialization predicted
by the prior is therefore
\begin{equation}
\boldsymbol{u}
=
\begin{bmatrix}
\boldsymbol{w}_1^{\top} &
\cdots &
\boldsymbol{w}_5^{\top} &
T_1 &
\cdots &
T_6
\end{bmatrix}^{\top}
\in\mathbb{R}^{36}.
\label{eq:prior_trajectory_representation}
\end{equation}
The first $30$ entries describe the five intermediate whole-body waypoints, and
the remaining six entries describe the segment durations.

\begin{figure}[t]
    \centering
    \includegraphics[width=\linewidth]
    {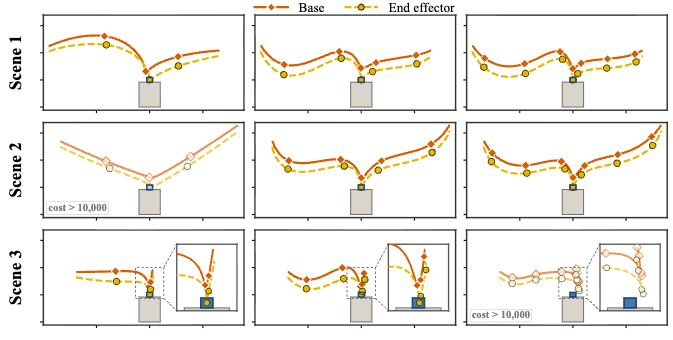}
    \caption{Selection of the number of intermediate waypoints for the trajectory prior. The columns compare trajectories represented by three, five, and seven intermediate waypoints under three representative boundary configurations. Solid orange and dashed yellow curves denote the aerial platform and end-effector trajectories, respectively. Diamond and circular markers indicate their corresponding intermediate waypoints. Faded trajectories exceed the prescribed objective threshold.}
    \label{fig:waypoint_number_selection}
\end{figure}

The number of intermediate waypoints is selected by balancing trajectory representation capability and optimization complexity, as illustrated in \Cref{fig:waypoint_number_selection}. Three intermediate waypoints provide insufficient flexibility for complex motions, whereas seven introduce unnecessary optimization variables and closely spaced segments. Five intermediate waypoints achieve a suitable compromise across the representative boundary configurations and are therefore used throughout the learning framework.

Because the position variables and segment durations have different numerical scales, they are normalized before learning. Each waypoint coordinate is normalized independently using its mean and standard deviation in the training set:
\begin{subequations}
\label{eq:trajectory_normalization}
\begin{align}
\widehat{u}_{k}
&=
\frac{u_k-\mu_k}{\sigma_k},
\qquad
k\in\{1,\ldots,30\},
\label{eq:spatial_variable_normalization}
\\
\widehat{u}_{30+i}
&=
\frac{\log T_i-\mu_{T,i}}{\sigma_{T,i}},
\qquad
i\in\{1,\ldots,6\},
\label{eq:duration_normalization}
\end{align}
\end{subequations}
where $\mu_k$ and $\sigma_k$ are the mean and standard deviation of the
corresponding waypoint coordinate, while $\mu_{T,i}$ and $\sigma_{T,i}$ are computed from the logarithmic segment durations. We denote the normalization and its inverse by $\mathcal{N}_{z}$ and $\mathcal{N}_{z}^{-1}$, respectively.

The trajectory prior is implemented as a multilayer perceptron that maps the local boundary condition to the normalized initialization:
\begin{equation}
\widehat{\boldsymbol{u}}
=
f_{\theta}
\left(
\boldsymbol{c}
\right),
\qquad
f_{\theta}:
\mathbb{R}^{6}
\rightarrow
\mathbb{R}^{36}.
\label{eq:trajectory_prior_mapping}
\end{equation}
All waypoint and duration variables are predicted jointly. The corresponding
initialization in physical units is recovered as
\begin{equation}
\boldsymbol{u}
=
\mathcal{N}_{z}^{-1}
\left(
\widehat{\boldsymbol{u}}
\right).
\label{eq:prior_seed_prediction}
\end{equation}

The denormalized prediction $\boldsymbol{u}$ is used as the initialization of
the trajectory optimizer, which refines it to obtain
\begin{equation}
\boldsymbol{u}^{\star}
=
\mathcal{O}
\left(
\boldsymbol{u};
\boldsymbol{c}
\right).
\label{eq:prior_initialized_optimization}
\end{equation}

\subsection{Offline Supervised Learning}
\label{subsec:offline_supervised_learning}

Without a learned prior, each grasping task is initialized by Heuristic Initialization, which combines path searching with heuristic time allocation to generate a trajectory seed. 
These procedures provide only a coarse initial guess; the resulting trajectory seed is not guaranteed to satisfy the complete trajectory requirements.
The optimizer may therefore spend substantial computation correcting the heuristic waypoint and duration assignment and reducing the associated geometric, temporal, kinematic, dynamic and grasp-specific penalties 
before obtaining a numerically acceptable whole-body trajectory.

Such repeated rediscovery of trajectory structures is unnecessary because aerial grasping trajectories share recurring spatial and temporal patterns across different boundary conditions. 
Offline Supervised Learning uses previously optimized trajectories to learn these patterns, including the coordination between the aerial platform and the manipulator, the typical arrangement of intermediate waypoints, and the allocation of segment durations. 
The resulting prior can therefore provide a structured initialization directly, rather than reconstructing the same grasping motion from a heuristic seed for every new task.

For the $i$-th scene sampled in the local range, where $i\in\{1,\ldots,N_{\mathrm{off}}\}$, let $\boldsymbol{c}_i$ denote the start and end position defined in \Cref{eq:local_prior_condition}. 
Heuristic Initialization uses the front-end path-search algorithm and SFC construction
to generate an initial trajectory seed:
\begin{equation}
\boldsymbol{u}_{i}^{\mathrm{h}}
=
\mathcal{H}
\left(
\boldsymbol{c}_{i}
\right),
\label{eq:heuristic_initialization}
\end{equation}
where $\mathcal{H}(\cdot)$ denotes the complete Heuristic Initialization
procedure. The trajectory optimizer subsequently refines the initial seed:
\begin{equation}
\boldsymbol{u}_{i}^{\star}
=
\mathcal{O}
\left(
\boldsymbol{u}_{i}^{\mathrm{h}};
\boldsymbol{c}_{i}
\right),
\label{eq:offline_trajectory_refinement}
\end{equation}
where $\boldsymbol{u}_{i}^{\star}$ denotes the optimized 36-dimensional trajectory variable defined in \Cref{eq:trajectory_parameter}.

The resulting offline supervision dataset is represented as:
\begin{equation}
\mathcal{D}_{\mathrm{off}}
=
\left\{
\left(
\boldsymbol{c}_{i},
\boldsymbol{u}_{i}^{\star}
\right)
\right\}_{i=1}^{N_{\mathrm{off}}}.
\label{eq:offline_supervision_dataset}
\end{equation}
The prior is trained by minimizing
\begin{equation}
\theta_{\mathrm{off}}
=
\arg\min_{\theta}
\frac{1}{N_{\mathrm{off}}}
\sum_{i=1}^{N_{\mathrm{off}}}
\left\|
f_{\theta}
\left(
\boldsymbol{c}_{i}
\right)
-
\mathcal{N}_{z}
\left(
\boldsymbol{u}_{i}^{\star}
\right)
\right\|_{2}^{2}.
\label{eq:offline_supervision_loss}
\end{equation}

The learned prior replaces Heuristic Initialization by directly providing the initialization to the trajectory optimizer, while the subsequent optimization process remains unchanged. Since it is trained as a deterministic regressor on a fixed offline dataset, its prediction reflects the representative, and potentially averaged, solution structure contained in the available optimized trajectories. The resulting initialization may therefore remain suboptimal.

\subsection{Online Evolution}
\label{subsec:online_evolution}

Offline Supervised Learning captures the representative solution structure contained in the fixed offline dataset. However, its deterministic prediction may still provide a suboptimal initialization for the deployed trajectory optimizer. 
Online Evolution therefore starts from the Offline checkpoint, $\theta_0=\theta_{\mathrm{off}}$, and further adapts the trajectory prior using direct optimizer feedback.

Online Evolution adapts the sampling and elite-selection principle of the CEM to conditional trajectory initialization. In standard CEM, candidate solutions are sampled from a parameterized proposal distribution and evaluated using a performance function. The best-performing candidates are retained as elites, and the proposal parameters are updated by maximizing the likelihood of these elites:
\begin{equation}
\boldsymbol{\eta}_{t+1}
=
\operatorname*{arg\,max}_{\boldsymbol{\eta}}
\sum_{\boldsymbol{x}\in\mathcal{E}_{t}}
\log
g
\left(
\boldsymbol{x};
\boldsymbol{\eta}
\right),
\label{eq:cem_elite_update}
\end{equation}
where $g(\boldsymbol{x};\boldsymbol{\eta})$ denotes the sampling distribution
and $\mathcal{E}_{t}$ contains the selected elite samples. 
If $g$ is Gaussian and its covariance is fixed during the current collection round, maximizing \Cref{eq:cem_elite_update} updates only its mean.

We adapt this principle to evolve a conditional trajectory prior.
Let $\theta_t$ denote the prior parameters before the $t$-th replay update. 
These parameters remain fixed while a batch of new conditions is processed. For each condition $\boldsymbol{c}_i$, the current prior output
\begin{equation}
\boldsymbol{\mu}_{i,t}
=
f_{\theta_t}
\left(
\boldsymbol{c}_i
\right)
\label{eq:online_sampling_mean}
\end{equation}
defines the mean of the sampling distribution in the normalized trajectory
space. We retain this mean as the nominal candidate and sample the remaining $K_t-1$ candidates from a Gaussian distribution:
\begin{subequations}
\label{eq:online_candidate_generation}
\begin{align}
\widehat{\boldsymbol{u}}_{i,1}
&=
\boldsymbol{\mu}_{i,t},
\\
\widehat{\boldsymbol{u}}_{i,j}
&\sim
\mathcal{N}
\left(
\boldsymbol{\mu}_{i,t},
\boldsymbol{\Sigma}_{t}
\right),
\qquad
j=2,\ldots,K_t.
\end{align}
\end{subequations}
Here, the learned prior determines the proposal mean $\boldsymbol{\mu}_{i,t}$, while $\boldsymbol{\Sigma}_{t}$ is a prescribed covariance that controls the exploration range.
The candidate number $K_t$ and covariance $\boldsymbol{\Sigma}_{t}$ are gradually reduced according to a fixed schedule, providing broader exploration in the early rounds and more concentrated sampling around the evolved prior prediction later.

Each normalized candidate is transformed back to the trajectory representation and independently refined by the same trajectory optimizer:
\begin{subequations}
\label{eq:online_candidate_evaluation}
\begin{align}
\boldsymbol{u}_{i,j}
&=
\mathcal{N}_{z}^{-1}
\left(
\widehat{\boldsymbol{u}}_{i,j}
\right),
\\
\boldsymbol{u}_{i,j}^{\star}
&=
\mathcal{O}
\left(
\boldsymbol{u}_{i,j};
\boldsymbol{c}_i
\right).
\end{align}
\end{subequations}

For each candidate, we record the resulting numerical objective
$J_{i,j}
=
\widehat{\mathcal{J}}
(\boldsymbol{u}_{i,j}^{\star};\boldsymbol{c}_i)$
and optimizer iteration count $n_{i,j}$.
The objective value measures the quality of the refined trajectory, while the iteration count reflects the additional solver effort required from the initialization.

Candidates satisfying the objective threshold form the retained set
\begin{equation}
\mathcal{F}_{i}
=
\left\{
j\in\{1,\ldots,K_t\}
\;\middle|\;
J_{i,j}
\leq
\tau_{J}^{\mathrm{target}}
\right\},
\label{eq:online_candidate_set}
\end{equation}
where $\tau_{J}^{\mathrm{target}}$ is the prescribed value of the numerical objective. If $\mathcal{F}_{i}$ is empty, the current condition does not contribute an elite target.

If $\mathcal{F}_i$ is nonempty, we retain one elite for condition $\boldsymbol{c}_i$: the candidate whose optimized trajectory has the lowest numerical objective. Its optimizer input is used as the supervision target for updating the trajectory prior:
\begin{equation}
\begin{aligned}
j_i^{\star}
&=
\operatorname*{arg\,min}_{j\in\mathcal{F}_{i}}
J_{i,j},
\\
\boldsymbol{u}_{i}^{\dagger}
&=
\boldsymbol{u}_{i,j_i^{\star}}.
\end{aligned}
\label{eq:online_candidate_selection}
\end{equation}
Here, $\boldsymbol{u}_{i}^{\dagger}$ denotes the optimizer input currently retained as the supervision target for condition $\boldsymbol{c}_i$. If the
selected candidate already requires fewer than
$\tau_{\mathrm{iter}}^{\mathrm{target}}$ optimizer iterations, this input is retained
directly.

If the selected candidate still requires excessive optimization iterations, additional refinement is performed by restarting the optimizer from the resulting optimized trajectory. 
The refined optimizer input replaces the supervision target if it achieves the desired objective within the iteration budget:
Let $\boldsymbol{u}_{i}^{(0)}=\boldsymbol{u}_{i,j_i^\star}$ and
$\boldsymbol{u}_{i}^{\star,(0)}
=\boldsymbol{u}_{i,j_i^\star}^{\star}$.
The subsequent calls are defined by
\begin{subequations}
\label{eq:online_reoptimization}
\begin{align}
\boldsymbol{u}_{i}^{(r+1)}
&=
\boldsymbol{u}_{i}^{\star,(r)},
\\
\boldsymbol{u}_{i}^{\star,(r+1)}
&=
\mathcal{O}
\left(
\boldsymbol{u}_{i}^{(r+1)};
\boldsymbol{c}_i
\right),
\end{align}
\end{subequations}
where each call uses the same cost function but
starts with a newly initialized optimizer.

For each call, we record the numerical objective
$J_i^{(r)}
=
\widehat{\mathcal{J}}
(\boldsymbol{u}_{i}^{\star,(r)};\boldsymbol{c}_i)$
and the iteration count $n_i^{(r)}$.  
Because each subsequent optimizer call starts from the optimized result of the preceding call, repeated optimization further refines the selected candidate under the same objective.
Since each refinement starts from the previous optimized trajectory, the objective values of successive calls are expected to be non-increasing.

Whenever the result of a new call remains below the target objective
$\tau_{J}^{\mathrm{target}}$, the input of that call replaces the currently retained supervision target:
\begin{equation}
\boldsymbol{u}_{i}^{\dagger}
\leftarrow
\boldsymbol{u}_{i}^{(r+1)}.
\label{eq:online_target_update}
\end{equation}
The repeated optimization terminates when the current call requires fewer than
$\tau_{\mathrm{iter}}^{\mathrm{target}}$ iterations or when the prescribed maximum number of calls is reached. Consequently, $\boldsymbol{u}_{i}^{\dagger}$ is the first optimizer input whose output satisfies the target objective within the desired iteration count. 

This refinement step accounts for the fact that additional solver iterations may reveal whether the selected initialization is close to a favorable optimization basin.


Overall, Online Evolution starts from the prior obtained through Offline Supervised Learning and treats its prediction as the mean of a conditional Gaussian proposal. 
For each condition, it samples candidate initializations, evaluates them using the deployed optimizer, and retains the optimizer input associated with the best refined candidate as an elite. 
Because the current prior prediction is always included, every elite is selected against the nominal initialization. 
Repeating this CEM-based sampling, optimizer evaluation, elite selection, and regression update evolves the conditional mean represented by the Online prior toward initializations that are more favorable to the optimizer, while the covariance follows the prescribed exploration schedule.

\subsection{Execution-Aware Critic}
\label{subsec:execution_aware_critic}

\begin{figure}[t]
    \centering
    \includegraphics[width=1.0\linewidth]{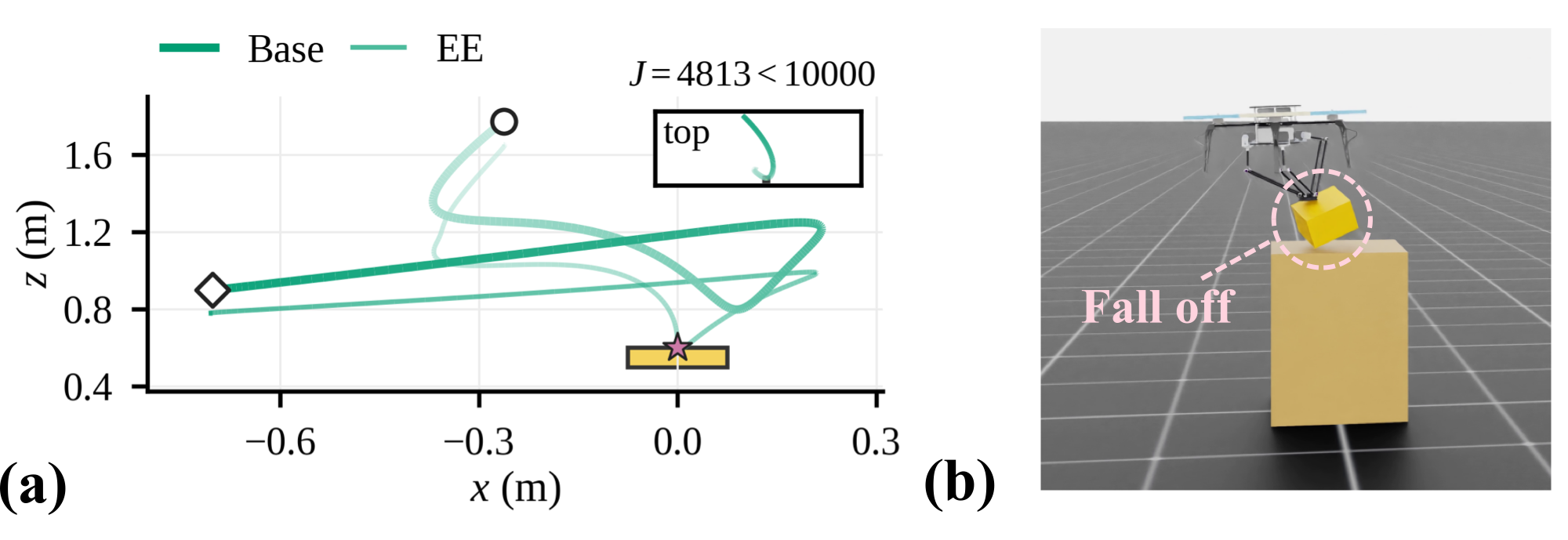}
   \caption{Example showing that satisfying the objective threshold does not guarantee successful execution. (a) The optimized trajectory satisfies the objective threshold $J<\tau_{J}^{\mathrm{target}}$. (b) The object nevertheless falls from the end effector during closed-loop execution.}
    \label{fig:objective_execution_gap}
\end{figure}

Online Evolution improves the trajectory initialization according to the numerical objective used by the trajectory optimizer. 
However, as discussed in \Cref{subsec:motion_ambiguity}, a low numerical objective does not necessarily imply successful task execution. 
A trajectory may satisfy the prescribed grasp requirements while still failing due to physical interaction effects that are not captured by the numerical objective.
Therefore,
\begin{equation}
J_{i,j}
\leq
\tau_{J,\mathrm{target}}
\;\nRightarrow\;
y_{i,j}^{\mathrm{complete}}
=
1,
\label{eq:objective_execution_gap}
\end{equation}
where $y_{i,j}^{\mathrm{complete}}$ indicates whether candidate $j$ in scene $i$ completes the entire grasping task. As illustrated in \Cref{fig:objective_execution_gap}, a trajectory can satisfy the objective threshold while failing to retain the object during execution.

Rather than replacing the analytical planner, we retain its existing geometric, temporal, kinematic and dynamic constraints and learn the execution preference that is not fully represented by the numerical objective.
After Online Evolution, the trajectory prior is frozen. 
We then train an EAC to evaluate complete optimized trajectories using physical execution outcomes. 
EAC therefore learns an execution-aligned preference over a fixed, structured trajectory representation.

\subsubsection{Trajectory Features}

EAC evaluates each optimized candidate using a fixed
feature vector. For candidate $j$ under condition
$\boldsymbol{c}_i$, let $T_{i,j}$ denote the total trajectory duration and
$t_{i,j}^{g}$ denote the time of the designated grasp waypoint. We additionally
select a pre-grasp time $t_{i,j}^{\mathrm{pre}}$ and a post-grasp time
$t_{i,j}^{\mathrm{post}}$ around $t_{i,j}^{g}$.

Let
$\boldsymbol{\varphi}_{i,j}(t)\in\mathbb{R}^{15}$
denote the fixed motion descriptor extracted from the optimized continuous
trajectory at time $t$. 
It contains the selected position, velocity, and acceleration quantities of the aerial platform and the relative end-effector motion, resulting in a 15-dimensional descriptor.
We define the three local descriptors as
\begin{equation}
\begin{aligned}
\boldsymbol{\varphi}_{i,j}^{\mathrm{pre}}
&=
\boldsymbol{\varphi}_{i,j}
\left(
t_{i,j}^{\mathrm{pre}}
\right),\\
\boldsymbol{\varphi}_{i,j}^{g}
&=
\boldsymbol{\varphi}_{i,j}
\left(
t_{i,j}^{g}
\right),\\
\boldsymbol{\varphi}_{i,j}^{\mathrm{post}}
&=
\boldsymbol{\varphi}_{i,j}
\left(
t_{i,j}^{\mathrm{post}}
\right).
\end{aligned}
\label{eq:eac_local_features}
\end{equation}

We additionally include
\begin{equation}
\boldsymbol{\eta}_{i,j}
=
\begin{bmatrix}
\log T_{i,j} &
\bar{t}_{i,j}^{g} &
b_{i,j}^{g}
\end{bmatrix}^{\top}
\in\mathbb{R}^{3},
\label{eq:eac_temporal_features}
\end{equation}
where $\bar{t}_{i,j}^{g}$ and $b_{i,j}^{g}$ denote the normalized grasp time and the distance between the optimized and the desired grasp point.

The complete EAC input is constructed by the fixed feature mapping
\begin{equation}
\begin{aligned}
\boldsymbol{\xi}_{i,j}
=
\begin{bmatrix}
\boldsymbol{c}_{i}\\
\boldsymbol{u}_{i,j}^{\star}\\
\boldsymbol{\eta}_{i,j}\\
\boldsymbol{\varphi}_{i,j}^{\mathrm{pre}}\\
\boldsymbol{\varphi}_{i,j}^{g}\\
\boldsymbol{\varphi}_{i,j}^{\mathrm{post}}
\end{bmatrix}
=
\Psi
\left(
\boldsymbol{c}_{i},
\boldsymbol{u}_{i,j}^{\star}
\right)
\in\mathbb{R}^{90}.
\end{aligned}
\label{eq:eac_feature}
\end{equation}

The mapping $\Psi$ is fixed before training, while EAC learns the nonlinear relationship between the resulting trajectory features and the observed execution outcomes.

\subsubsection{Execution Supervision}

Each optimized trajectory is evaluated using three binary execution checks. The
first check determines whether the intended contact is fully established, the
second determines whether the object is successfully lifted, and the third
determines whether the entire grasping task is completed. The resulting labels
are
\begin{equation}
\boldsymbol{y}_{i,j}
=
\begin{bmatrix}
y_{i,j}^{\mathrm{contact}} &
y_{i,j}^{\mathrm{lift}} &
y_{i,j}^{\mathrm{complete}}
\end{bmatrix}^{\top}
\in
\{0,1\}^{3}.
\label{eq:eac_labels}
\end{equation}
These labels are obtained from separate checks during execution.

The EAC maps the trajectory features to three corresponding logits:
\begin{equation}
\boldsymbol{\ell}_{i,j}
=
h_{\phi}
\left(
\boldsymbol{\xi}_{i,j}
\right)
=
\begin{bmatrix}
\ell_{i,j}^{\mathrm{contact}} &
\ell_{i,j}^{\mathrm{lift}} &
\ell_{i,j}^{\mathrm{complete}}
\end{bmatrix}^{\top}.
\label{eq:eac_logits}
\end{equation}
The pointwise binary cross-entropy loss supervises the three execution checks
independently.

To distinguish trajectories that have similar numerical objectives but different
task outcomes, we additionally construct successful--failed pairs from the same
scene. Let $j^{+}$ denote a trajectory that completes the entire task and
$j^{-}$ denote a failed trajectory from the same scene. Their completion logits
are trained using the Bradley--Terry loss
\begin{equation}
\mathcal{L}_{\mathrm{pair}}
=
\mathbb{E}_{(i,j^{+},j^{-})}
\left[
\operatorname{softplus}
\left(
-
\left[
\ell_{i,j^{+}}^{\mathrm{complete}}
-
\ell_{i,j^{-}}^{\mathrm{complete}}
\right]
\right)
\right].
\label{eq:eac_pairwise_loss}
\end{equation}

Although the three outcomes are evaluated separately, they follow a natural hierarchy:
successful completion requires successful lifting, and successful lifting requires successful contact.
We therefore introduce a soft monotonicity regularizer to enforce this ordering by penalizing the lift logit exceeding the contact logit and the completion logit exceeding the lift logit.
The complete training objective is
\begin{equation}
\mathcal{L}_{\mathrm{EAC}}
=
\mathcal{L}_{\mathrm{BCE}}
+
\mathcal{L}_{\mathrm{pair}}
+
0.1\mathcal{L}_{\mathrm{mono}}.
\label{eq:eac_training_loss}
\end{equation}

\subsubsection{Conservative Execution Score}

We train an ensemble of three critics. 
For each execution check
$h\in
\{\mathrm{contact},\mathrm{lift},\mathrm{complete}\}$,
let $\mu_h$ and $\sigma_h$ denote the mean and standard deviation of the logits predicted by the ensemble models.
The conservative logit is
\begin{equation}
\widetilde{\ell}_{h}
=
\mu_h
-
\beta
\sigma_h,
\label{eq:eac_conservative_logit}
\end{equation}
where $\beta$ controls the conservativeness induced by ensemble disagreement.

For each execution check, the conservative logit is compared with a prescribed
target margin. 
If the logit exceeds the margin, the corresponding penalty is close to zero; otherwise, the penalty increases smoothly as the logit falls below the margin.

The penalties from the three execution checks are combined as
\begin{equation}
E_{\mathrm{EAC}}
=
\sum_{h}
w_h T
\operatorname{softplus}
\left(
\frac{
m_h-\widetilde{\ell}_{h}
}{T}
\right),
\label{eq:eac_energy}
\end{equation}
where $m_h$ is the target margin for execution check $h$, $w_h$ is its
weight, and $T$ controls the smoothness of the penalty. A lower
$E_{\mathrm{EAC}}$ indicates that the trajectory is more likely to satisfy all
three execution checks. Ensemble disagreement lowers
$\widetilde{\ell}_{h}$ and therefore increases the energy assigned to an
uncertain trajectory.

For convenience, we define the EAC score as the negative energy:
\begin{equation}
s_{\mathrm{EAC}}
=
-
E_{\mathrm{EAC}}.
\label{eq:eac_score}
\end{equation}
A higher $s_{\mathrm{EAC}}$ indicates that the trajectory is predicted to be
more likely to complete the entire grasping task successfully. 
The EAC score is evaluated after trajectory optimization, and only trajectories
that reach the prescribed threshold are admitted for execution.

\subsection{Deployment}
\label{subsec:deployment_generation}

Using this decision rule, deployment proceeds sequentially. The Online prior
first predicts the nominal initialization, which is refined by the original
trajectory optimizer. The optimized trajectory is accepted if
$J_i^{(0)}\leq\tau_{J}^{\mathrm{target}}$ and
$s_{\mathrm{EAC},i}^{(0)}
\geq\tau_{\mathrm{EAC}}^{\mathrm{target}}$.

If the current trajectory is not accepted, the next initialization is generated
by adding the prescribed perturbation to the original prediction.
This initialization is independently refined by the same optimizer and checked
using the same objective and EAC criteria.
The procedure terminates when a trajectory is accepted or when the maximum number
of four optimization attempts, including the nominal initialization, is reached.
If no trajectory reaches the acceptance threshold, trajectory generation is considered unsuccessful.

The Online prior therefore provides the initializations, the trajectory optimizer produces the candidate trajectories, and EAC provides the execution-aware acceptance decision.

\section{Experiments}
\label{sec:experiments}

We evaluate the proposed framework through six questions.

\textbf{Q1)} How do Offline Supervised Learning, Online Evolution, and EAC contribute to planning and execution performance?

\textbf{Q2)} Does the EAC improvement come from additional candidate coverage or from using predicted physical execution outcomes?

\textbf{Q3)} Where should execution feedback be introduced in the proposed framework?

\textbf{Q4)} What trajectory family does Online Evolution produce?

\textbf{Q5)} What does EAC evaluate in an optimized grasping trajectory, and can this learned preference be used directly for trajectory optimization?

\textbf{Q6)} Do the resulting improvements transfer to real-world aerial grasping?

We first describe the common experimental setup and then address these questions in order.

\subsection{Experimental Setup}
\label{subsec:experimental_setup}

\subsubsection{Hardware Platform}
\label{subsubsec:hardware_platform}

The aerial manipulator used in the real-world experiments is shown in
 \Cref{fig:first_grasp}. It consists of a quadrotor equipped with a three-DoF delta arm. The onboard system includes a computer with an Intel Processor N150, an NxtPX4v2 flight controller, and three DYNAMIXEL
XL430-W250-T servo motors. Robot states are estimated using an EKF that fuses measurements from a NOKOV motion-capture system and the onboard IMU. 
All simulation-based data collection and network training are conducted on a
workstation equipped with an Intel Xeon Platinum 8370C CPU at $2.80\,\mathrm{GHz}$ and a single NVIDIA GeForce RTX 4090 GPU with $24\,\mathrm{GB}$ memory.
Trajectory optimization is executed on the CPU, whereas the GPU is used for parallel simulation and network training.

\subsubsection{Simulation Setup}
\label{subsubsec:software_simulation_setup}

\begin{figure}[t]
    \centering
    \includegraphics[width=\linewidth]{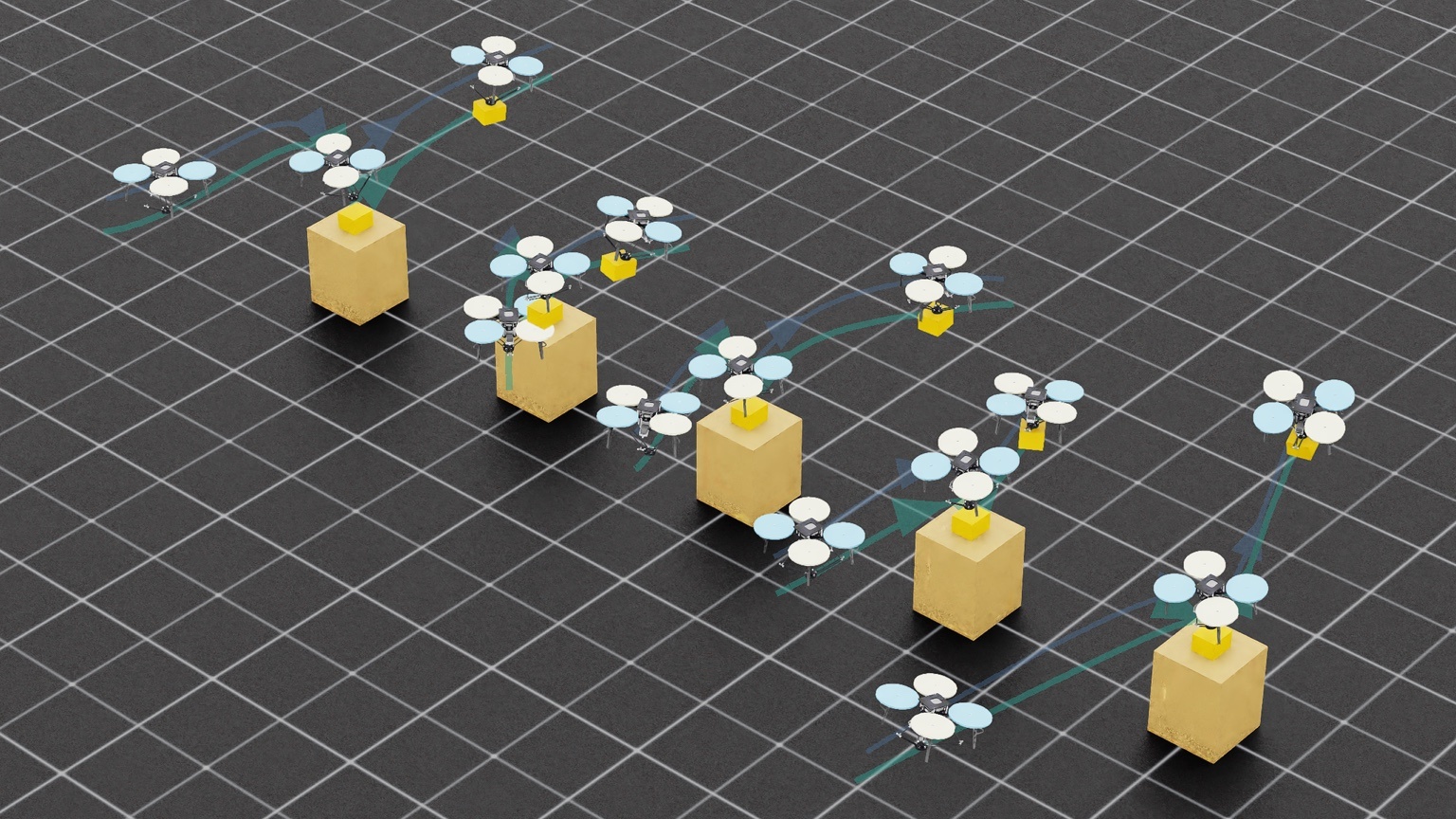}
    \caption{GPU-parallel closed-loop aerial-grasping simulation. Multiple planned trajectories are executed concurrently in vectorized Isaac Sim environments.}
    \label{fig:parallel_simulation}
\end{figure}

The planning and control system is implemented in C++ under Ubuntu 20.04 and ROS Noetic. We developed a deterministic, GPU-parallel closed-loop execution simulator using Isaac Sim 4.5.0, Isaac Lab 2.1.0, and PhysX. Multiple planned trajectories are executed concurrently in vectorized simulation environments, as illustrated in \Cref{fig:parallel_simulation}.

The simulator samples the aerial platform and manipulator references at $100\,\mathrm{Hz}$ and tracks them using a cascaded geometric controller consisting of position, attitude, and body-rate control. The resulting thrust and torque commands are converted into individual rotor commands through the physical allocation matrix, quadratic thrust model, and first-order motor response before the resulting wrench is applied to the multi-link PhysX model. Motor delay, saturation, trajectory tracking error, manipulator motion, gravity, contact, and payload reaction can therefore affect the execution outcome. All compared methods use the same controller, physical parameters, and execution criteria, so their performance differences arise from the initialization and trajectory-evaluation procedures being compared.

\subsubsection{Training}

Offline Supervised Learning and Online Evolution share the same trajectory prior network, a $6$-$256$-$512$-$256$-$36$ MLP with $273{,}956$ trainable parameters and a dropout rate of $0.05$. Offline Supervised Learning uses $5{,}000$ trajectories obtained by initializing the planner with path search and SFC construction and retaining the corresponding trajectories that satisfy the numerical objective threshold. The prior is trained with Adam using a learning rate of $3\times10^{-4}$, weight decay of $10^{-6}$, and a batch size of $256$. Starting from the Offline checkpoint, Online Evolution retains $12{,}000$ selected optimizer inputs as supervision. Online updates use a replay buffer with a batch size of $512$, learning rate of $3\times10^{-5}$, and weight decay of $10^{-6}$.

EAC uses an ensemble of three independently trained models. Each model has a $90$-$64$-$32$-$3$ architecture with LayerNorm, SiLU activation, and a dropout rate of $0.1$ after each hidden layer, resulting in $8{,}195$ parameters per model. The EAC is trained on $227{,}626$ labeled trajectories. Each EAC model is trained with AdamW using a learning rate of $3\times10^{-4}$, weight decay of $10^{-5}$, and a batch size of $2{,}048$. Training runs for at most $50$ epochs with an early-stopping patience of $8$, and gradients are clipped at $5.0$. The training objective follows \Cref{eq:eac_training_loss}, with unit weights for the three execution heads and the pairwise term and a weight of $0.1$ for the monotonicity regularizer.

\subsection{Planning and Execution Performance}
\label{subsec:end_to_end_performance}

This experiment examines how Offline Supervised Learning, Online Evolution, and EAC contribute to planning and execution performance while the trajectory optimizer and execution controller remain unchanged. We compare Heuristic Initialization, Offline Supervised Learning, Online Evolution, and EAC on 5,000 paired evaluation scenes disjoint from the training and calibration data. Heuristic Initialization uses path search and SFC construction, whereas Offline and Online each provide one prior prediction followed by one optimization. EAC starts from the Online Evolution prediction and introduces another initialization only when the current optimized trajectory is rejected, with at most four optimization attempts. A scene is successful only if the trajectory establishes contact, lifts the object, and retains it until the end of execution. All numerical and physical failures remain in the denominator.

\begin{figure*}[t]
    \centering
    \includegraphics[width=\textwidth]{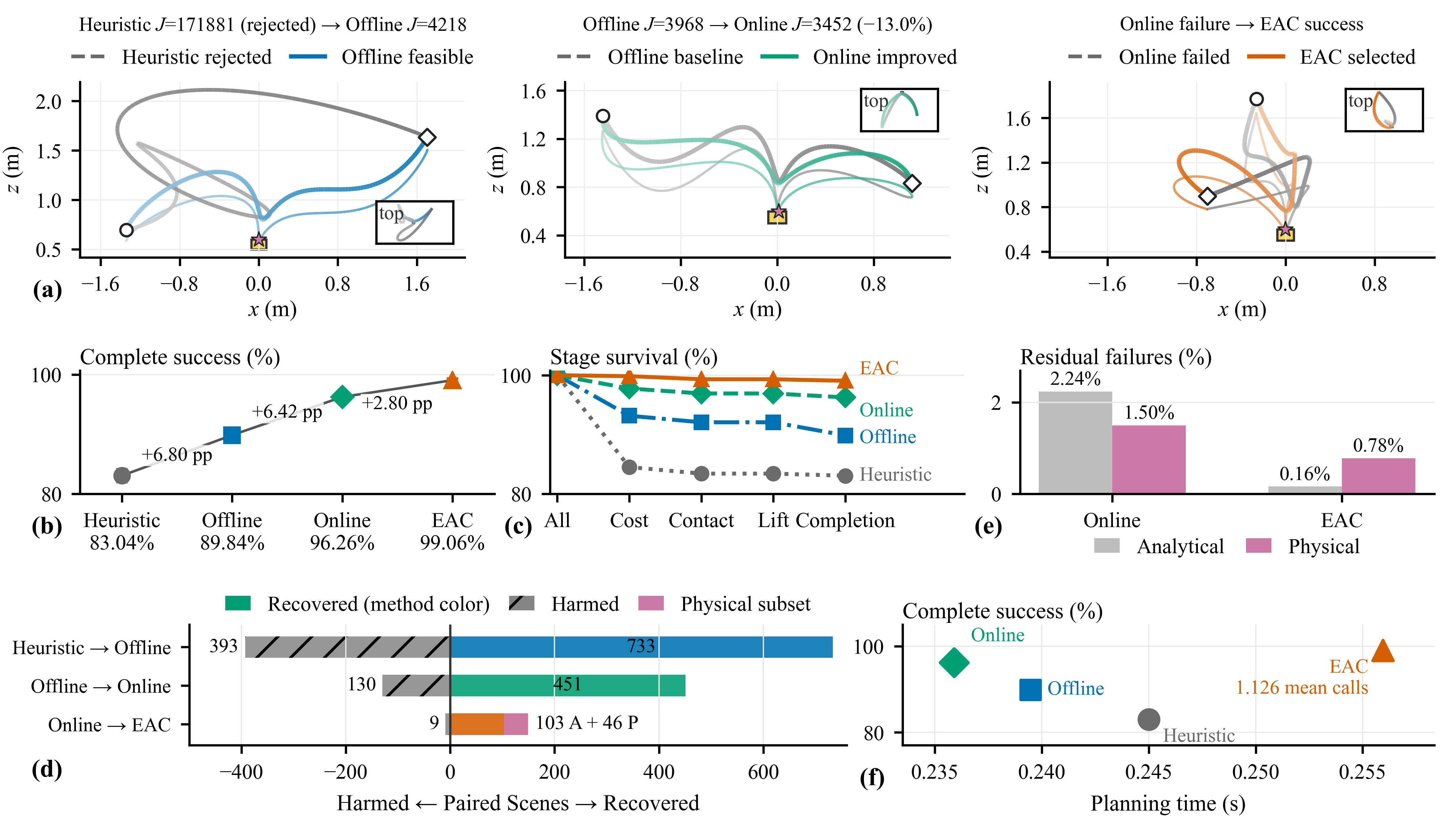}
   \caption{Planning and execution performance on 5,000 paired scenes. (a) Representative improvements from Offline Supervised Learning, Online Evolution, and EAC. (b) Complete success. (c) Fraction of scenes passing the numerical objective, contact, lift, and completion checks. (d) Recovered and harmed scenes between adjacent methods. (e) Remaining analytical and physical failures after Online Evolution and EAC. (f) Complete success versus p90 planning time.}
    \label{fig:progressive_results}
\end{figure*}

The representative cases in \Cref{fig:progressive_results} (a) illustrate the different contributions of the three learned components. Offline Supervised Learning replaces a heuristic initialization whose optimized result exceeds the numerical objective threshold, Online Evolution further reduces the objective relative to Offline Supervised Learning, and EAC recovers an Online trajectory that fails during physical execution. As shown in \Cref{fig:progressive_results} (b), complete success increases from $83.04\%$ with Heuristic Initialization to $89.84\%$ with Offline Supervised Learning and $96.26\%$ with Online Evolution, corresponding to gains of $6.80$ and $6.42$ percentage points, respectively. 
Importantly, the gains from Offline Supervised Learning and Online Evolution arise mainly from improving the outcome of trajectory optimization. As shown in \Cref{fig:progressive_results} (c), the fraction of scenes for which the optimizer produces a trajectory satisfying the numerical objective increases from $84.50\%$ with Heuristic Initialization to $93.16\%$ with Offline Supervised Learning and $97.76\%$ with Online Evolution. 
These results indicate that Offline Supervised Learning provides more effective initializations than the heuristic procedure, while Online Evolution further adapts the prior using feedback from the deployed optimizer, allowing acceptable solutions to be reached more reliably.

EAC further increases complete success from $96.26\%$ to $99.06\%$. Relative to Online Evolution, it recovers 149 failed scenes while 9 previously successful scenes become unsuccessful, yielding a net gain of $2.80$ percentage points. 
\Cref{fig:progressive_results} (d) separates the 149 recoveries into 103 cases in which an additional initialization produces a trajectory satisfying the numerical objective and 46 cases in which an analytically qualified Online trajectory fails during execution but another EAC-evaluated trajectory succeeds. 
This gain comes from two sources: trying additional initializations and using EAC to assess the optimized trajectories. The second effect is examined separately in \Cref{subsec:candidate_coverage_eac}.

As shown in \Cref{fig:progressive_results} (e), Online Evolution leaves 112 analytical failures and 75 physical failures, whereas the EAC deployment reduces them to 8 and 39, respectively. The total number of failures decreases from 187 to 47, a reduction of $74.9\%$. This improvement is obtained without applying the full optimization budget to every scene. The nominal Online initialization is sufficient in 4,691 of the 5,000 scenes ($93.82\%$), resulting in only $1.1258$ optimizer calls on average. 
As shown in \Cref{fig:progressive_results} (f), the p90 planning time increases from $235.9$ to $255.9\,\mathrm{ms}$ relative to Online Evolution, while complete success increases by $2.80$ percentage points. 
Overall, Offline Supervised Learning provides more effective initializations, Online Evolution further improves them using solver feedback, and EAC uses execution feedback to determine whether the optimized result should be accepted or another initialization should be evaluated.

\subsection{Candidate Coverage and EAC Selection}
\label{subsec:candidate_coverage_eac}
During deployment, EAC evaluates each optimized trajectory and triggers another initialization and optimization when the current trajectory is rejected, resulting in an average of $1.13$ optimizer calls per scene. The improvement in execution success may therefore come from additional candidate coverage or EAC-based evaluation. We design the following comparisons to separate these two contributions.

We evaluate a fresh set of 5,000 scenes using five selection strategies.
As shown in \Cref{tab:candidate_coverage}, Online mean optimizes only the initialization predicted by Online Evolution and therefore uses one optimizer call. 
Cost-first K4 additionally optimizes three perturbed initializations and selects the trajectory with the lowest numerical objective among the four candidates. 
EAC fixed K4 uses exactly the same four optimized candidates but selects the one with the highest EAC score. 
EAC deployed follows the actual deployment strategy: EAC evaluates each optimized trajectory and triggers another initialization and optimization when the current trajectory is rejected. 
Finally, Oracle K4 selects the physically successful trajectory whenever one exists among the same four candidates, providing an upper bound for this candidate set.

\begin{table}[t]
    \centering
    \caption{Candidate coverage and selection results over 5,000 scenes.}
    \label{tab:candidate_coverage}
    \footnotesize
    \setlength{\tabcolsep}{3pt}
    \begin{tabular*}{\columnwidth}{
        @{\extracolsep{\fill}}
        l
        c
        c
        c
        @{}
    }
        \toprule
        Method &
        Success (\%) &
        \shortstack{Mean calls} &
        \shortstack{Opt. time (ms)} \\
        \midrule
        Online mean    & 95.74          & 1.00          & 55.25  \\
        Cost-first K4  & 98.64          & 4.00          & 240.01 \\
        EAC fixed K4   & \textbf{99.34} & 4.00          & 240.01 \\
        EAC deployed   & 99.18          & 1.13          & 64.08  \\
        Oracle K4      & 99.60          & 4.00          & 240.01 \\
        \bottomrule
    \end{tabular*}
\end{table}

Cost-first K4 increases the success rate from $95.74\%$ for Online mean to $98.64\%$, a gain of $2.90$ percentage points. Since both methods select trajectories only according to the numerical objective, this gain quantifies the benefit of additional candidate coverage provided by the three perturbed initializations. EAC fixed K4 selects from exactly the same four optimized trajectories and increases the success rate further to $99.34\%$. The additional $0.70$ percentage-point gain therefore measures the contribution of EAC-based selection under the same candidate set and optimization budget. Oracle K4 reaches $99.60\%$, leaving a $0.26$ percentage-point gap between EAC selection and the best achievable result from these four candidates.

EAC deployed achieves $99.18\%$ success with only $1.13$ optimizer calls and $64.08\,\mathrm{ms}$ of mean planning time. Compared with EAC fixed K4, it reduces the average number of calls from $4.00$ to $1.13$ with only a $0.16$ percentage-point decrease in success. 
Compared with Online mean, the additional $0.13$ calls per scene increase success by $3.44$ percentage points. 

These results show that additional initializations improve candidate coverage, EAC provides further gains when selecting among the same candidates, and the deployed sequential strategy retains most of this improvement with a small increase in optimizer calls.

\subsection{Ablation of Execution Feedback}
\label{subsec:execution_feedback_placement}
\begin{table}[t]
    \centering
    \caption{Results of four designs for incorporating execution feedback over 5,000 evaluation scenes.}
    \label{tab:execution-feedback-ablation}
    \footnotesize
    \setlength{\tabcolsep}{3pt}
    \begin{tabular*}{\columnwidth}{
        @{\extracolsep{\fill}}
        l
        c
        c
        c
        @{}
    }
        \toprule
        Design &
        Success (\%) &
        \shortstack{$\Delta$ Success (pp)} &
        \shortstack{Mean calls} \\
        \midrule
        GSD  & 95.92          & +0.14          & 1.00  \\
        RIR  & 95.78          & +0.00          & 1.00 \\
        EGJE & 93.94          & -1.84          & 1.00\\
        \textbf{EAC}
             & \textbf{99.18}
             & \textbf{+3.40}
             & 1.13 \\
        \bottomrule
    \end{tabular*}
\end{table}

\begin{figure}[t]
    \centering
    \includegraphics[width=1.0\linewidth]{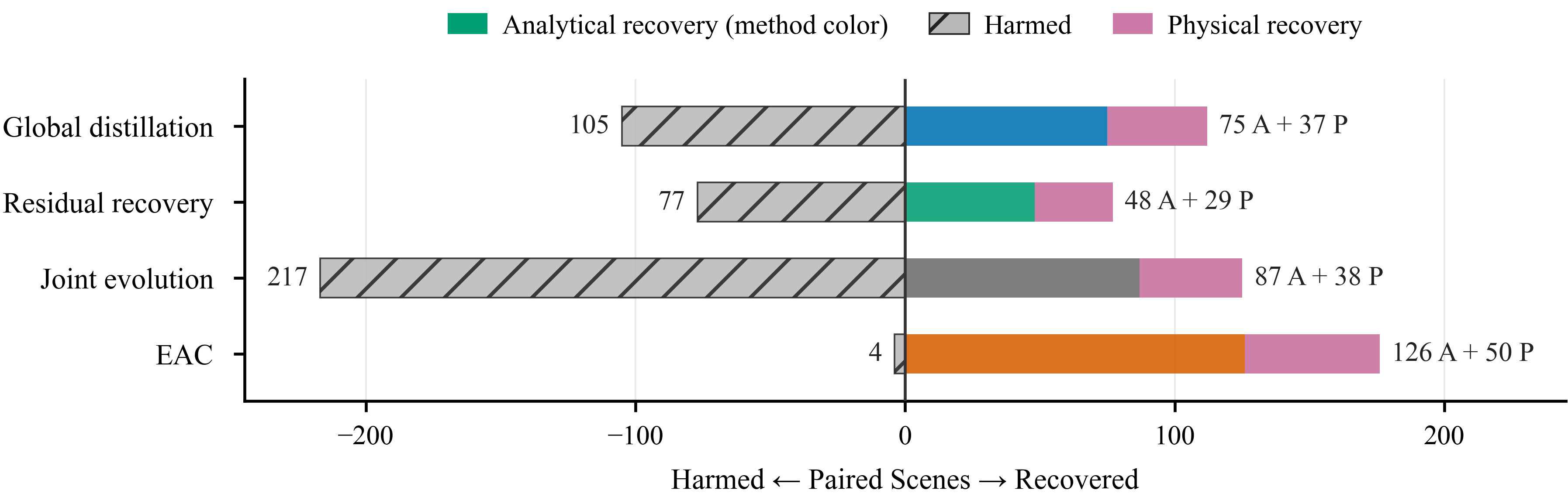}
    \caption{Paired recoveries and harms for the four execution-feedback designs. Bars to the left show previously successful scenes that become unsuccessful, while bars to the right show analytical and physical recoveries. GSD, RIR, and EGJE are paired with the same Online Evolution reference, while EAC is paired with the Online mean used in its deployed candidate generation.}
    \label{fig:execution_feedback_ablation}
\end{figure}

Execution outcomes provide direct supervision of physical task success, which raises an important design question: should this feedback be incorporated into the prior during Online Evolution, or applied after trajectory optimization through EAC? We investigate this question by evaluating three alternative designs that introduce execution outcomes into prior learning, together with the adopted EAC design.

\textit{Global Solution Distillation} (GSD) starts from the frozen Online prior and continues the sample-optimize-replay procedure used in Online Evolution. 
Replay samples are additionally required to produce physically successful optimized trajectories, and the selected optimizer inputs are used to finetune the prior. 
\textit{Residual Initialization Recovery} (RIR) also finetunes the Online prior, focusing specifically on conditions for which the nominal Online trajectory satisfies the analytical requirements but fails during execution; successful alternative initializations found for these cases are used for further training. 
\textit{Execution-Gated Joint Evolution} (EGJE) introduces execution outcomes earlier, starting from the Offline checkpoint and requiring physical success in addition to the original numerical criteria throughout Online Evolution. 
EAC leaves the learned prior unchanged and evaluates complete optimized trajectories during deployment, generating another candidate when the current result is rejected.

All four designs are evaluated over 5,000 fresh scenes. GSD, RIR, and EGJE retain one predicted initialization and one optimizer call at deployment, since execution feedback has already been incorporated into the prior. 
EAC follows the deployed sequential strategy and introduces another initialization only when the current optimized trajectory is rejected, resulting in $1.13$ optimizer calls on average.

As shown in \Cref{tab:execution-feedback-ablation} and \Cref{fig:execution_feedback_ablation}, incorporating execution outcomes into prior learning provides little or negative overall improvement. 
GSD recovers 112 failures but harms 105 previously successful scenes, producing only a $0.14$ percentage-point gain. 
RIR produces the same number of recoveries and harms and leaves the success rate unchanged. 
EGJE recovers 125 failures but harms 217 successful scenes, reducing success by $1.84$ percentage points.

These results show the difficulty of using execution outcomes to update a prior shared across the complete task-condition distribution. 
GSD and RIR can correct some failed conditions, but finetuning the prior also changes its predictions for conditions that already produce successful trajectories. 
EGJE applies the execution requirement throughout Online Evolution and removes many samples that satisfy the numerical optimization criteria, which reduces the supervision available for learning useful optimizer initializations. The resulting recoveries are therefore offset by failures introduced at other conditions.

EAC avoids changing the prior learned from optimizer feedback and applies the execution outcome to the optimized trajectory for the current condition. It recovers 126 analytical and 50 physical failures while harming only four successful scenes, reaching $99.18\%$ success with an average of $1.13$ optimizer calls. Among the evaluated designs, this is the only placement of execution feedback that produces a clear improvement over the full evaluation set.

Overall, this experiment explains the separation between Online Evolution and EAC in the proposed framework. Online Evolution uses numerical optimizer feedback to learn initializations across the task-condition distribution, while EAC uses execution outcomes after optimization to evaluate the trajectory generated for the current condition. Among the evaluated designs, this arrangement achieves the largest improvement in physical success while preserving the prior learned from optimizer feedback.

\begin{figure*}[t]
    \centering
    \includegraphics[width=\textwidth]{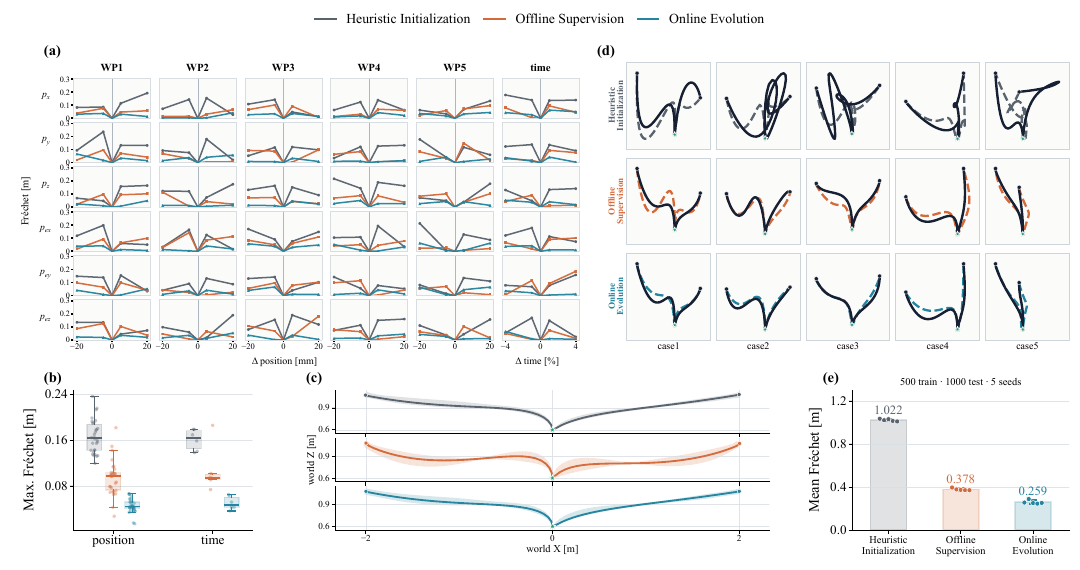}
    \caption{Trajectory consistency after Heuristic Initialization, Offline Supervised Learning, and Online Evolution.
(a) Three-dimensional Fr\'echet distance between final trajectories from perturbed and unperturbed initializations; the first five columns perturb the five intermediate waypoints and the last perturbs segment durations.
(b) Maximum responses to the position and duration perturbations in (a).
(c) World-frame $x$-$z$ medoid trajectories under local perturbations of the initial and terminal positions, with shaded 10th-90th percentile normal displacements shown at twice their actual width for visibility.
(d) Optimized trajectories (solid) and trajectories reconstructed from a task-conditioned MLP (dashed) for five representative conditions.
(e) Three-dimensional Fr\'echet reconstruction errors of the MLP across unseen task conditions. All Fr\'echet distances are computed in three-dimensional world space.}
    \label{fig:trajectory_consistency}
\end{figure*}

\subsection{Trajectory Consistency after Online Evolution}
\label{subsec:trajectory_consistency}

This experiment examines whether Online Evolution only reduces the numerical objective or also organizes the optimized trajectories into a more consistent family of grasping motions. 
We test this from three aspects: sensitivity to initialization perturbations, consistency under nearby task conditions, and regularity across unseen conditions. 
We compare Heuristic Initialization, Offline Supervised Learning, and Online Evolution to isolate the effect of evolving the trajectory prior.
Spatial differences are measured using the three-dimensional discrete Fr\'echet distance over 65 equally spaced arc-length samples.

\subsubsection{Consistency under Initialization Perturbations}

\Cref{fig:trajectory_consistency} (a) measures the sensitivity of the optimized trajectory to perturbations of the initialization. 
The first five columns correspond to the five intermediate waypoints, and the six rows correspond to the platform and end-effector coordinates $p_x$, $p_y$, $p_z$, ${}^{D}p_{ex}$, ${}^{D}p_{ey}$, and ${}^{D}p_{ez}$. 
For these waypoint variables, the horizontal axis gives the applied position perturbation from $-20$ to $20\,\mathrm{mm}$. 
The final column instead perturbs the segment durations, with the horizontal axis showing the relative time perturbation. 
In all panels, the vertical axis is the three-dimensional Fr\'echet distance between the trajectories optimized from the perturbed and original initializations, so a smaller value indicates lower sensitivity to the initialization. 
A clear overall ordering is observed across most perturbations: Online Evolution produces the smallest trajectory deviations, followed by Offline Supervised Learning and Heuristic Initialization. \Cref{fig:trajectory_consistency} (b) further summarizes the maximum trajectory deviation under the waypoint and duration perturbations. Online Evolution maintains small deviations for both types of perturbations, showing that the optimized trajectories remain highly consistent even when the initialization is perturbed.

\subsubsection{Consistency under Nearby Conditions}

\Cref{fig:trajectory_consistency} (c) evaluates the effect of locally perturbing the initial and terminal positions. The horizontal and vertical axes show the world-frame $x$ and $z$ coordinates, the solid curve denotes the medoid trajectory, and the shaded region shows the 10th to 90th percentiles of the displacement normal to the medoid. 
Heuristic Initialization and Online Evolution both produce narrow trajectory distributions, whereas Offline Supervised Learning exhibits substantially larger variation. 
For Heuristic Initialization, the limited variation mainly arises because small changes in the initial and terminal positions often leave the searched path and resulting SFC unchanged, leading to similar heuristic initializations despite their higher numerical objective. 
Offline Supervised Learning learns lower-cost grasping structures from the fixed offline dataset, but exhibits greater trajectory variation across nearby task conditions, suggesting that small condition changes may lead the optimizer to different local solutions.
Online Evolution further updates the prior using feedback from the deployed optimizer, reducing this variation across nearby task conditions. As a result, the optimized trajectories become more consistent under local changes in the task condition while maintaining a low numerical objective.

\subsubsection{Consistency across Different Conditions}
We next examine whether optimized trajectories obtained under different start and end conditions exhibit regular variations associated with the task conditions. 
To evaluate this regularity, we use a compact neural network to learn the mapping from task conditions to optimized trajectories. 
For each method, we train an independent MLP with an identical $6$-$32$-$36$ architecture.
The input contains the initial and terminal positions, while the output consists of 30 waypoint coordinates for the aerial platform and arm, together with six segment durations. These parameters are converted into continuous trajectories using the MINCO formulation and flatness mapping.

Each model is trained on 500 task conditions and evaluated on 1,000 unseen conditions, with five random seeds. All three methods use identical network architectures, training settings, and task conditions. This controlled comparison allows us to assess how regularly the optimized trajectories vary with the task conditions.

\Cref{fig:trajectory_consistency} (d) compares the optimized trajectories with the MLP reconstructions under five representative task conditions. The solid curves denote the optimized trajectories, while the dashed curves denote the reconstructed trajectories. Heuristic Initialization exhibits substantial discrepancies, which are reduced by Offline Supervised Learning. Online Evolution shows the closest agreement between the optimized and reconstructed trajectories, suggesting more regular variations across different task conditions.

\Cref{fig:trajectory_consistency} (e) quantitatively evaluates the reconstruction errors using the three-dimensional discrete Fr\'echet distance over 65 equally spaced arc-length samples. The mean errors are $1.022\ \mathrm{m}$ for Heuristic Initialization, $0.378\ \mathrm{m}$ for Offline Supervised Learning, and $0.259\ \mathrm{m}$ for Online Evolution, where the standard deviations are computed across five random seeds. Online Evolution reduces the reconstruction error by $31.5\%$ compared with Offline Supervised Learning under the same model capacity and training conditions. These results suggest that Online Evolution organizes optimized trajectories into a more regular family of grasping motions, whose variations are more systematically related to task conditions.

\subsection{From Execution Assessment to Grasping Cost}
\label{subsec:execution_aligned_cost}

EAC has so far been used after trajectory optimization to assess whether an optimized trajectory is likely to succeed, without modifying the trajectory optimization process itself. This raises a further question: can the same learned execution preference be used directly in trajectory optimization? We examine this without retraining EAC. Specifically, we remove the prescribed grasp position, velocity, and orientation penalties and use the frozen EAC energy as the grasping cost, while retaining the remaining analytical planning terms.

Both formulations start from the same Online prediction and retain the same geometric, temporal, kinematic and dynamic penalties. Let $J_{\mathrm{generic}}$ denote this shared planning cost. The complete Position-Velocity-Orientation (PVO) cost is
\begin{subequations}
\label{eq:pvo_grasping_cost}
\begin{align}
J_{\mathrm{PVO}}(\boldsymbol{z})
&=
J_{\mathrm{generic}}(\boldsymbol{z})
+
J_{\mathrm{grasp}}^{\mathrm{PVO}}(\boldsymbol{z}),
\\
J_{\mathrm{grasp}}^{\mathrm{PVO}}(\boldsymbol{z})
&=
\lambda_e J_{pe}(\boldsymbol{z})
+
\lambda_v J_{ve}(\boldsymbol{z})
\nonumber\\
&\quad
+
\lambda_o J_{oe}(\boldsymbol{z}).
\end{align}
\end{subequations}

For the EAC formulation, the prescribed PVO penalties are removed. The fixed feature mapping $\Psi$ maps the task condition and trajectory to $\boldsymbol{\xi}=\Psi(\boldsymbol{c},\boldsymbol{z})$. The complete cost becomes
\begin{subequations}
\label{eq:eac_grasping_cost}
\begin{align}
J_{\mathrm{EAC}}(\boldsymbol{z})
&=
J_{\mathrm{generic}}(\boldsymbol{z})
+
J_{\mathrm{grasp}}^{\mathrm{EAC}}
(\boldsymbol{c},\boldsymbol{z})
\nonumber\\
&\quad
+
\lambda_{\mathrm{support}}
J_{\mathrm{support}}(\boldsymbol{\xi}),
\\
J_{\mathrm{grasp}}^{\mathrm{EAC}}
(\boldsymbol{c},\boldsymbol{z})
&=
\lambda_{\mathrm{EAC}}
E_{\mathrm{EAC}}(\boldsymbol{\xi}).
\end{align}
\end{subequations}
Here, $\lambda_{\mathrm{EAC}}$ and $\lambda_{\mathrm{support}}$ are the corresponding weights. 
The support penalty $J_{\mathrm{support}}$ discourages the optimizer from driving the standardized EAC features toward or beyond six standard deviations from their corresponding means in the training data.
The EAC energy evaluates the task condition and planned trajectory features around grasping, including waypoints, segment durations, and platform and arm states. Importantly, it has no access to realized execution outcomes during optimization; contact, lifting, and object retention are never observed when evaluating a candidate trajectory.

\begin{figure}[t]
    \centering
    \includegraphics[width=\linewidth]{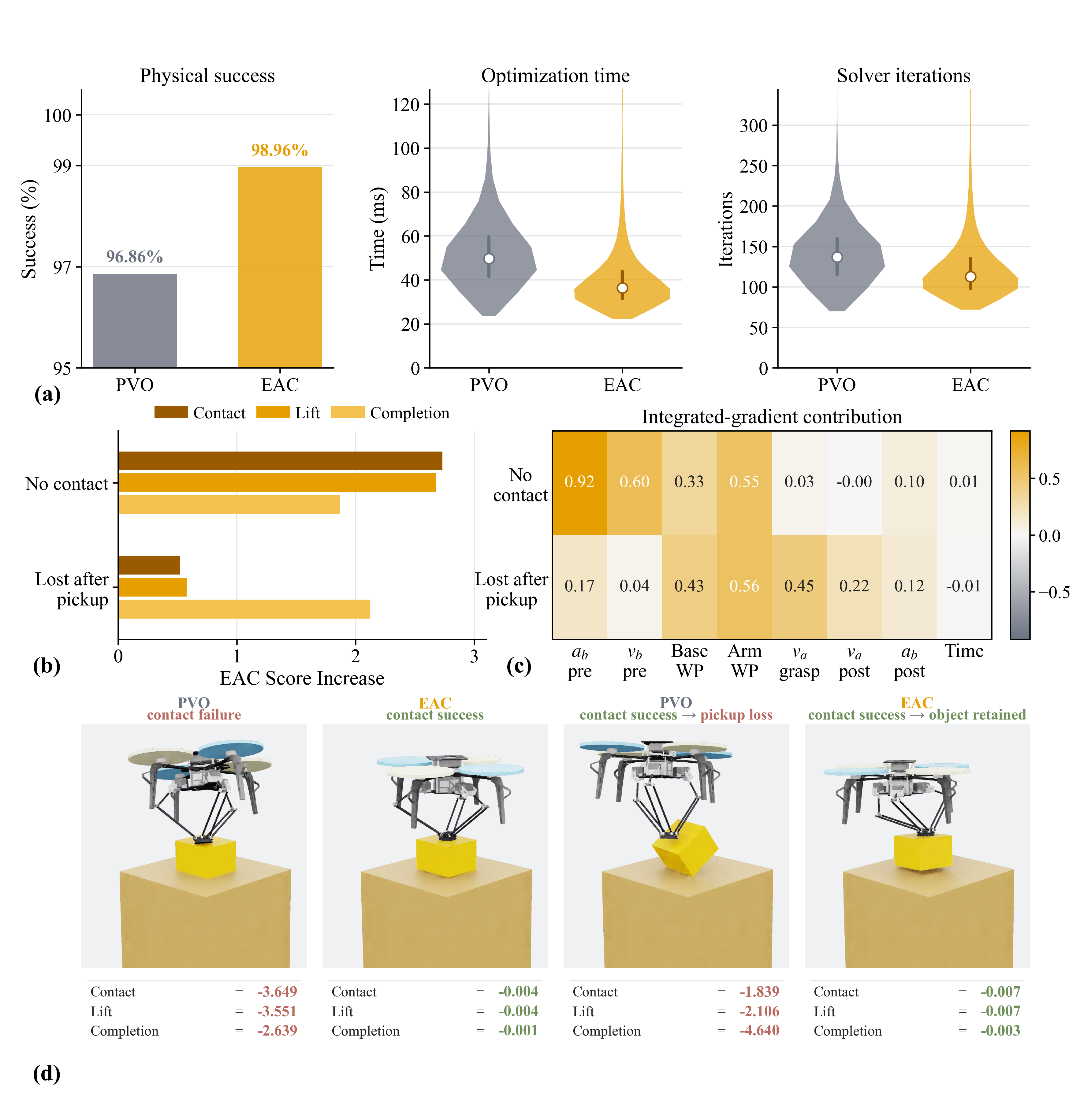}
    \caption{Comparison between the prescribed PVO grasping cost and the grasping cost learned from execution data. (a) Physical success, optimization time, and solver iterations. (b) Changes in the Contact, Lift, and Completion scores for No contact and Lost after pickup recoveries. (c) Integrated Gradients attribution of EAC score changes to motion features. (d) Representative execution snapshots for two recovery cases from the same initialization. The corresponding Contact, Lift, and Completion scores are shown below each execution.}
    \label{fig:execution_aligned_cost}
\end{figure}

\Cref{fig:execution_aligned_cost} (a) compares the two formulations across 5,000 paired task conditions. 
Replacing the prescribed PVO grasping cost with EAC increases physical success from $96.86\%$ to $98.96\%$. 
EAC recovers 124 PVO failures while introducing 19 new failures, yielding 105 additional successes and reducing the total number of failures from 157 to 52, or by $66.9\%$. 
Meanwhile, the median optimization time decreases from $49.79$ to $36.42\,\mathrm{ms}$, and the median solver iteration count decreases from 137 to 113.

To understand these recoveries, \Cref{fig:execution_aligned_cost} (b) examines the 124 cases in which \textbf{PVO fails but EAC succeeds}. These include 59 \textit{No contact} cases and 65 \textit{Lost after pickup} cases, classified according to the PVO execution outcomes. We evaluate both trajectories using the same frozen EAC and compare their Contact, Lift, and Completion scores.

For \textit{No contact} recoveries, the EAC-optimized trajectories receive substantially higher scores at all three stages. For \textit{Lost after pickup} recoveries, the largest improvement occurs in the Completion score, consistent with the improved object retention. These differences reveal distinct patterns in the learned execution scores that correspond to the stages at which the failures are recovered.

\Cref{fig:execution_aligned_cost} (c) further examines which motion features contribute to the improvement in the EAC Completion score. We use Integrated Gradients with 33 interpolation points between the PVO and EAC feature representations to attribute the score changes to individual input features.
For \textit{No contact} recoveries, the largest contributions arise from the aerial platform acceleration and velocity before grasping, together with the arm waypoints. Since acceleration is closely coupled to attitude for underactuated quadrotors, these features are relevant to the end-effector approach and alignment before contact.
For \textit{Lost after pickup} recoveries, the dominant contributions shift toward the arm waypoints and arm velocity at grasping, with additional contributions from the platform waypoint. These features are associated with the relative motion of the end-effector around pickup and the subsequent object retention.
These results suggest that EAC captures execution preferences involving coordinated platform and arm motion beyond the prescribed position, velocity, and orientation at the grasp waypoint.

\Cref{fig:execution_aligned_cost} (d) presents two representative paired examples from the same initialization, illustrating how changes in the learned scores and motion features correspond to the observed execution behavior.
In the first pair, the PVO trajectory approaches the object with a visibly misaligned platform configuration and fails to establish contact. 
Correspondingly, its Contact, Lift, and Completion scores are strongly negative ($-3.649$, $-3.551$, and $-2.639$). 
The EAC-optimized trajectory instead produces a better aligned pre-grasp approach and successfully establishes contact, with all three scores becoming close to zero ($-0.004$, $-0.004$, and $-0.001$). 
This observation is consistent with the strong contributions of pre-grasp platform acceleration and velocity in \Cref{fig:execution_aligned_cost} (c).

In the second pair, the PVO trajectory reaches the object but fails to retain it during pickup. Its Lift and Completion scores are consequently strongly negative, particularly the Completion score of $-4.640$. In contrast, the EAC-optimized trajectory maintains the grasp and retains the object after pickup, with the three scores approaching zero ($-0.007$, $-0.007$, and $-0.003$). This behavior is consistent with both the dominant Completion improvement in \Cref{fig:execution_aligned_cost} (b) and the increased contribution of end-effector motion around grasping in \Cref{fig:execution_aligned_cost} (c).

Overall, the attribution analysis reveals which trajectory features contribute most to the EAC scores, providing an interpretable view of its learned execution preferences. Furthermore, incorporating the frozen EAC as a differentiable critic enables these learned preferences to directly guide trajectory generation, improving grasping success beyond manually prescribed grasping objectives.

\begin{figure*}[t]
    \centering
    \includegraphics[width=0.95\textwidth]{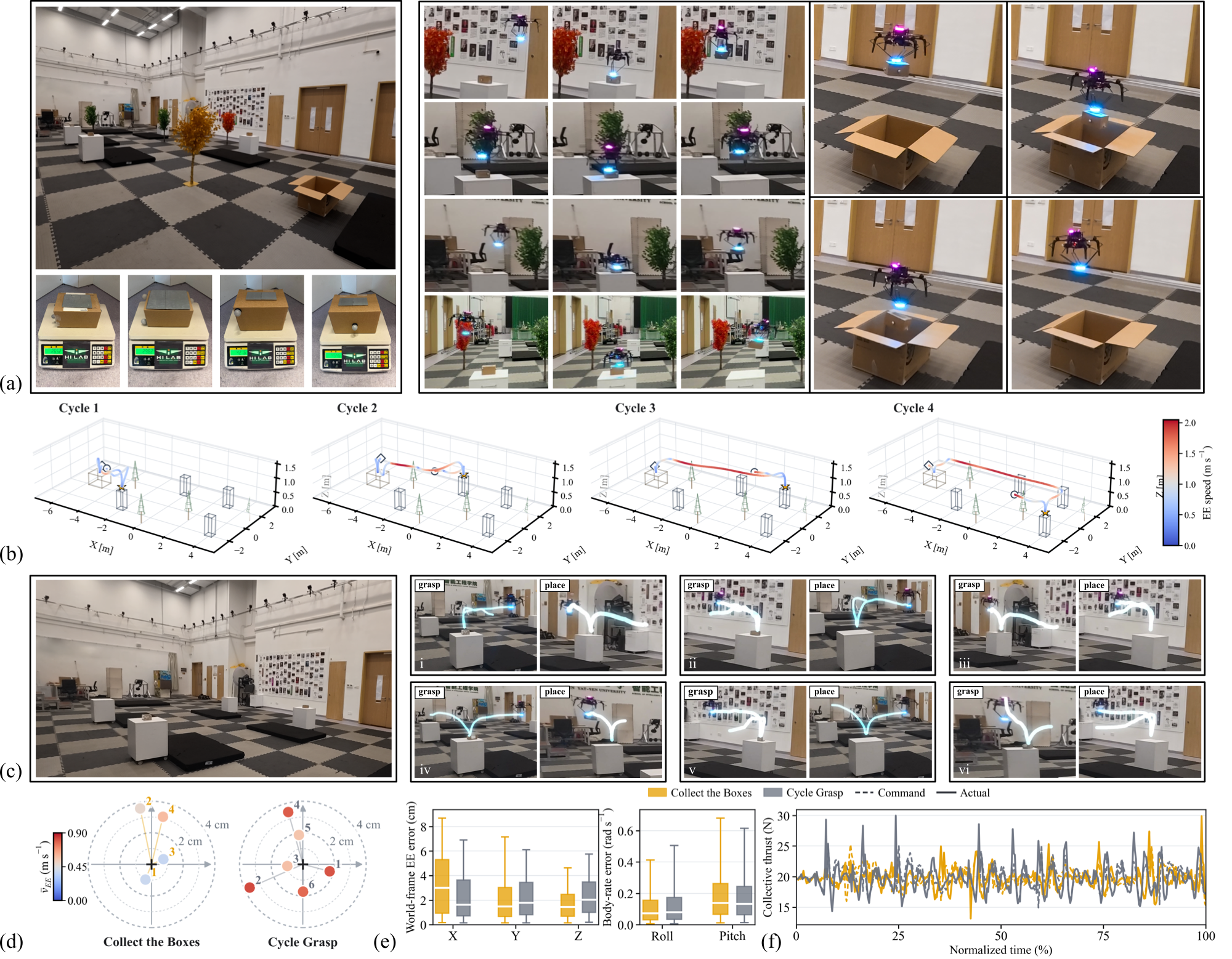}
    \caption{Real-world validation of our framework. 
    (a) Collect the Boxes scene, measured object masses, four grasp sequences, and delivery to the common collection box. 
    (b) Complete end-effector command trajectories for the four collection cycles in the world frame, with color indicating commanded end-effector speed. 
    (c) Cycle Grasp scene and six consecutive grasp-and-place pairs. 
    (d) Horizontal end-effector offsets relative to the object at the touch instant; marker color denotes the mean end-effector speed near grasping.
    (e) End-effector position tracking errors and roll and pitch body-rate tracking errors over the complete runs. 
    (f) Commanded and actual collective thrust over normalized execution time.}
    \label{fig:real_world_validation}
\end{figure*}

\subsection{Real-World Validation}
\label{subsec:real_world_grasping}

The preceding experiments evaluate the proposed components through paired simulation studies. 
We further examine whether the proposed framework can be integrated with the whole-body trajectory optimizer and flight controller for continuous real-world manipulation. 
Both experiments use the same aerial manipulator and control system described in \Cref{subsec:experimental_setup}.  
The prior predicts the waypoint and duration initialization, while the optimizer refines the complete trajectory under the same geometric, temporal, kinematic and dynamic constraints used throughout the framework. Retaining these analytical constraints allows the learned components to be directly deployed on the physical system without additional adaptation.
We consider two complementary tasks. \textit{Collect the Boxes} evaluates collection over a large workspace with obstacles and multiple objects, whereas \textit{Cycle Grasp} evaluates repeated planning and execution as the source and destination tables change.

Collect the Boxes is performed in a cluttered scene with four source tables, four tree obstacles, and one collection box, as shown in \Cref{fig:real_world_validation} (a). The four objects weigh $265.2$, $298.2$, $421.6$, and $454.6\,\mathrm{g}$. The aerial manipulator collects each object from a different table and places it in the shared box. 
The grasp and release sequences are shown in \Cref{fig:real_world_validation} (a). 
All four cycles, comprising eight manipulation tasks, are completed without failure in $68.84\,\mathrm{s}$. The corresponding end-effector command trajectories in \Cref{fig:real_world_validation} (b) cover different approach and transport paths through the same obstacle layout, with color indicating the commanded speed.

Cycle Grasp evaluates repeated planning as the object locations change. Two objects are initially placed on two of four tables. In each cycle, an object is moved from an occupied table to a randomly selected empty table, after which the table occupancy is updated. Only the end-effector contact point is specified for each grasp and placement, while the prior and optimizer generate the remaining motion. \Cref{fig:real_world_validation} (c) shows the six grasp and place pairs in execution order. All six cycles, comprising twelve manipulation tasks, are completed without failure in $72.48\,\mathrm{s}$.

As shown in \Cref{fig:real_world_validation} (d), the horizontal end-effector offsets for all ten grasps remain below $4\,\mathrm{cm}$, while the marker color denotes the mean end-effector speed around each grasp. \Cref{fig:real_world_validation} (e) summarizes the end-effector position and body-rate errors throughout both continuous runs. \Cref{fig:real_world_validation} (f) compares the commanded and actual collective thrust.

Together, the two runs demonstrate that our framework supports both multi-object collection in a cluttered environment and six consecutive object transfers, completing all ten cycles while maintaining grasp offsets below $4\,\mathrm{cm}$.

\section{Discussion}
\label{sec:discussion}

The proposed framework is not a fixed sequence of modules and is not limited to aerial grasping.
Online Evolution can be viewed more generally as a solver-feedback mechanism for nonconvex optimization: given a parameterized initialization, a repeatable local optimizer, and its objective and feasibility signals, the initialization model can be improved by sampling around its current prediction, evaluating the samples with the deployed solver, and learning from the selected solver inputs. 
EAC can be applied more generally to tasks with clear and observable event signals. When these events provide information about task success that is not fully captured by the numerical objective, they can be used to supervise a critic after optimization. Solver feedback therefore improves the initialization of nonconvex optimization, while event feedback evaluates the optimized result using task outcomes beyond the numerical objective.

\begin{figure}[t]
    \centering
    \includegraphics[width=0.85\linewidth]{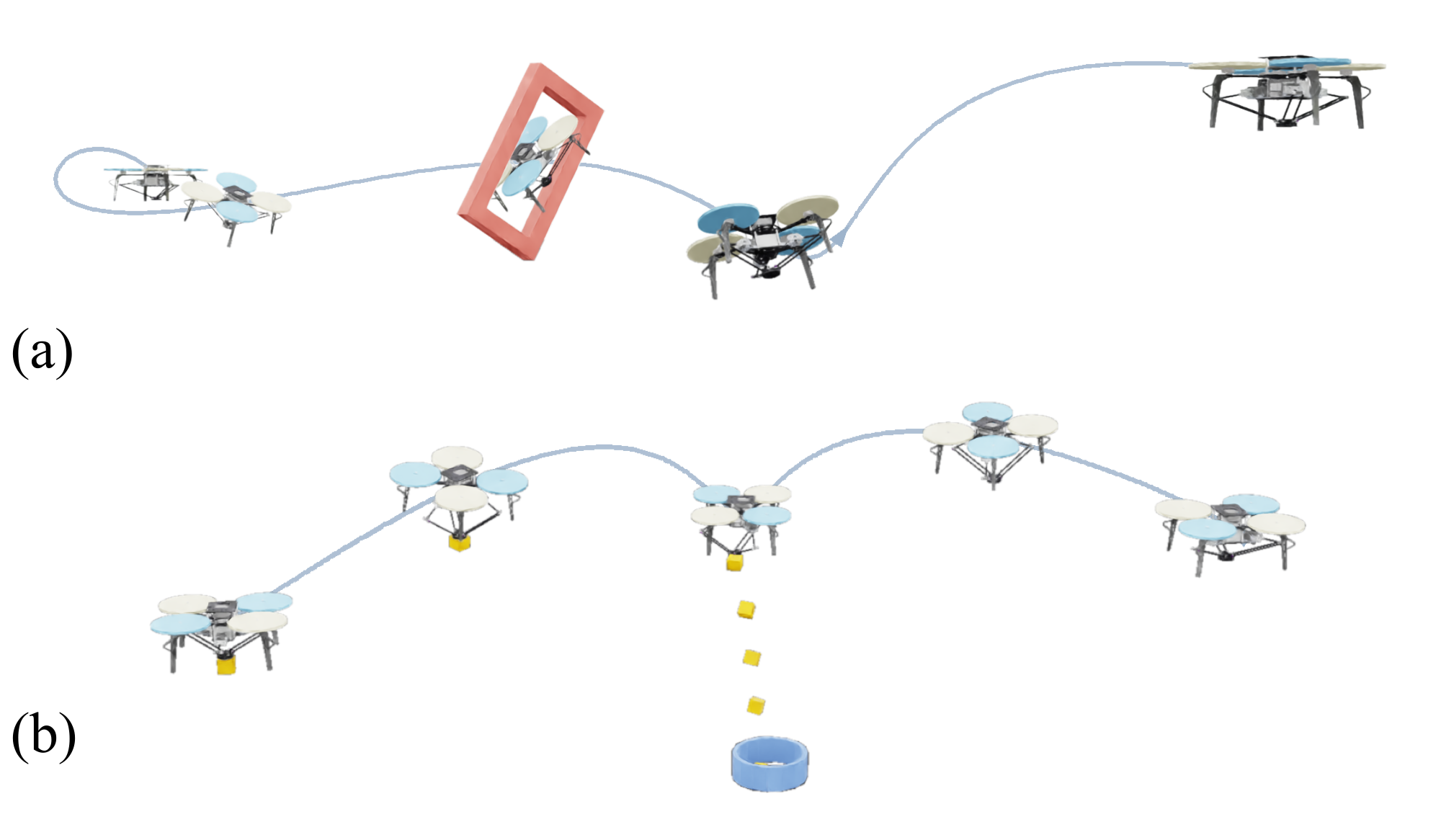}
    \caption{Cross-task examples of feedback placement. (a) Tilted Gate uses solver feedback to improve trajectory initialization for a recurring nonconvex maneuver. (b) Throw additionally uses release, full entry, and retention as execution supervision after trajectory optimization.}
    \label{fig:cross_task_transfer}
\end{figure}

\begin{table}[t]
    \centering
    \caption{Cross-task numerical comparison.}
    \label{tab:cross_task_transfer}
    \scriptsize
    \setlength{\tabcolsep}{2pt}
    \begin{tabular*}{\columnwidth}{
        @{\extracolsep{\fill}}
        lcccc
        @{}
    }
        \toprule
        & \multicolumn{2}{c}{Tilted Gate}
        & \multicolumn{2}{c}{Throw} \\
        \cmidrule(lr){2-3}\cmidrule(lr){4-5}
        Method & Cost ($\times 10^4$) & Iteration & Success (\%) & Iteration \\
        \midrule

        Heuristic
        & 5.33
        & 173.1
        & 56.10
        & 226.6 \\

        Offline
        & 2.44 {\scriptsize($\downarrow54.2\%$)}
        & 142.1 {\scriptsize($\downarrow17.9\%$)}
        & 77.80 {\scriptsize($\uparrow21.7~\mathrm{pp}$)}
        & 158.0 {\scriptsize($\downarrow30.3\%$)} \\

        Online
        & \textbf{2.35} {\scriptsize($\downarrow55.9\%$)}
        & \textbf{130.7} {\scriptsize($\downarrow24.5\%$)}
        & 77.20 {\scriptsize($\uparrow21.1~\mathrm{pp}$)}
        & 158.0 {\scriptsize($\downarrow30.3\%$)} \\

        EAC
        & --
        & --
        & \textbf{81.40} {\scriptsize($\uparrow25.3~\mathrm{pp}$)}
        & \textbf{146.1} {\scriptsize($\downarrow35.5\%$)} \\

        \bottomrule
    \end{tabular*}
\end{table}

To examine whether this principle extends beyond aerial grasping, we conduct two simple cross-task studies, as shown in \Cref{fig:cross_task_transfer} and \Cref{tab:cross_task_transfer}. 
Tilted Gate requires the aerial platform to traverse a narrow gate with varying orientation \cite{yang_whole-body_2021,wu_precise_2026}. 
Its numerical objective already provides a suitable measure of the desired maneuver, so solver feedback is sufficient: Offline Supervised Learning reduces the mean cost from $5.33\times10^{4}$ to $2.44\times10^{4}$, and Online Evolution further reduces it to $2.35\times10^{4}$ while also decreasing optimizer iterations. 
Throw instead requires a released payload to enter and remain inside a basket, making release, full entry, and retention explicit physical events that are not fully encoded by the numerical objective \cite{li_aerothrow_2025}. For throw, EAC further increases the success rate from $77.2\%$ to $81.4\%$, showing that execution feedback 
provides additional benefits for tasks with clear event signals.

The transfer is methodological. 
New tasks generally require task-specific condition and trajectory representations together with new Offline and Online training. 
EAC additionally requires reliable observable execution events. These studies illustrate that solver feedback can improve optimizer initialization, and execution feedback can provide additional supervision for tasks with clear event signals.

\section{Conclusion}
\label{sec:conclusion}
This work addressed two coupled limitations of aerial grasping planning: the sensitivity of nonconvex trajectory optimization to initialization and the misalignment between the numerical objective and physical task success. 
Offline Supervised Learning captures recurring whole-body motion structures from previously optimized trajectories, and Online Evolution further improves the predicted waypoint and duration initialization using feedback from the deployed optimizer. 
After the prior is frozen, the Execution-Aware Critic learns from contact, lift, and completion outcomes to evaluate optimized trajectories, triggering a new initialization from the prior only when the current result is rejected.

Together, these components raise complete success from $83.04\%$ with Heuristic Initialization to $96.26\%$ after Online Evolution and $99.06\%$ with EAC, while requiring only $1.13$ optimizer calls on average. Further comparisons show that EAC can reliably distinguish trajectories likely to succeed in physical execution and reject those likely to fail when necessary. 
Further analysis shows that Online Evolution produces optimized trajectories with reduced sensitivity to initialization perturbations, greater consistency under nearby task conditions, and more regular variations across different task conditions.

We further incorporate the frozen EAC energy as a differentiable grasping cost, replacing the prescribed position, velocity, and orientation penalties.
Feature attribution reveals which trajectory features contribute most to the EAC scores, providing insight into the execution preferences learned from physical outcomes.
As a differentiable critic, EAC enables these learned preferences to directly guide trajectory generation, capturing task requirements beyond those explicitly encoded in handcrafted grasping objectives.

Real-world demonstrations further show four object collections in a cluttered environment and six consecutive object transfers. 
Finally, we use cross-task studies to validate that this idea extends beyond aerial grasping: solver feedback can improve initialization for nonconvex optimization, and execution feedback can evaluate optimized solutions for tasks with clear event signals or even be integrated into optimization to directly shape trajectories.

\bibliographystyle{IEEEtran}
\bibliography{main}

\vfill

\end{document}